\documentclass[conference,compsoc]{IEEEtran}

\IEEEoverridecommandlockouts                               
\usepackage[table]{xcolor}
\usepackage{graphicx} 
\usepackage{chngpage}
\usepackage{calc}
\usepackage{epsfig} 
\usepackage{mathptmx} 
\usepackage{amsmath} 
\usepackage{amssymb}  
\usepackage{url}
\usepackage{enumitem}
\usepackage{multirow}
\usepackage{algorithm}
\usepackage{algorithmic}
\usepackage{subcaption}
\usepackage{xcolor}
\usepackage{xspace}
\usepackage{nicefrac}
\usepackage{numprint}
\usepackage{printlen}
\DeclareMathAlphabet{\mathcal}{OMS}{cmsy}{m}{n}
\newcommand{\ournameNoSpace}{\emph{BayBFed}}
\newcommand{\ourname}{\ournameNoSpace\xspace}
\newcommand{\ournameGen}{\ournameNoSpace's\xspace}
\newcommand{\adversaryNoSpace}{\ensuremath{\mathcal{A}}}
\newcommand{\adversary}{\adversaryNoSpace\xspace}

\newcommand{\fedavg}{\mbox{FedAVG}\xspace}
\newcommand{\diot}{\mbox{D\"IoT}\xspace}
\newcommand{\etal}{\emph{et~al.}}
\newcommand{\sect}{Sect.~}
\newcommand{\nonIidNoSpace}{non-IID}
\newcommand{\nonIid}{\nonIidNoSpace\xspace}
\newcommand{\iidNoSpace}{IID}
\newcommand{\iid}{\iidNoSpace\xspace}
\newcommand{\numberOfClients}{\ensuremath{n}}
\newcommand{\numberOfMaliciousClients}{\ensuremath{n_\adversary}\xspace}
\newcommand{\lnorm}{\ensuremath{L_2-\text{norm}}\xspace}

\newcommand{\sota}{state-of-the-art }
\newcommand{\jensenDivergence}{Jensen-Divergence\xspace}
\newcommand{\jensenDivergenceShort}{JD\xspace}

\newcommand{\ConstrainAndScale}{Constrain-and-Scale\xspace}

\definecolor{darkgreen}{rgb}{0.0, 0.5, 0.0}

\newcommand{\changed}[1]{#1}

\renewcommand{\changed}[1]{{\color{blue}{#1}}}
\definecolor{purple}{rgb}{0.5, 0.0, 0.5}

\newcommand{\scaleTable}[1]{\scalebox{0.925}{#1}}

\title{\LARGE \bf
\ourname: Bayesian Backdoor Defense for Federated Learning
}

\author{
\IEEEauthorblockN{Kavita Kumari${^\dagger}{^*}$, Phillip Rieger$^\dagger$, Hossein Fereidooni$^\dagger$, Murtuza Jadliwala$^\ddagger$ and Ahmad-Reza Sadeghi$^\dagger$}
\IEEEauthorblockA{\textit{$^\dagger$Technical University of Darmstadt, $^\ddagger$The University of Texas at San Antonio}\\
}

}
\begin{document}

\twocolumn
\setcounter{section}{0}
\setcounter{algorithm}{0}
\setcounter{figure}{0}
\maketitle
\thispagestyle{empty}
\pagestyle{empty}

\def\thefootnote{*}\footnotetext{Work done while author was affiliated with The University of Texas at San Antonio.}
\renewcommand*{\thefootnote}{\arabic{footnote}}

\renewcommand{\changed}[1]{#1}

\begin{abstract}
Federated learning (FL) is an emerging technology that allows 
participants to jointly train a machine learning model without sharing their private data with others.
However, FL is vulnerable to poisoning attacks such as backdoor attacks. 
Consequently, a variety of defenses \changed{have recently been} proposed, which have primarily utilized intermediary states of the global model (i.e., logits) or distance of the local models (i.e., \lnorm) with respect to the global model to detect malicious backdoors in FL. 
However, as these approaches \emph{directly} operate on client updates (or weights), their effectiveness depends on factors such as clients' data distribution or the adversary's attack strategies.  
\changed{In this paper, we introduce a novel and 
more generic backdoor defense framework, called \ourname, which proposes 
to utilize probability distributions over client updates to detect malicious updates in FL: 
\ourname computes a probabilistic measure over the clients' updates to keep track of any adjustments made in the updates, and uses a novel detection algorithm that can leverage this probabilistic measure to efficiently detect and filter out malicious updates. Thus, it overcomes the shortcomings of previous approaches that arise due to the \emph{direct} usage of client updates; nevertheless, our probabilistic measure will include all aspects of the local client training strategies.}  
\ourname utilizes two Bayesian Non-Parametric (BNP) extensions: (i) a Hierarchical Beta-Bernoulli process to draw a probabilistic measure given the clients' updates, and (ii) an adaptation of the Chinese Restaurant Process (CRP), referred by us as CRP-Jensen, which 
leverages this probabilistic measure to detect and filter out malicious updates.
We extensively evaluate our defense approach on \changed{five benchmark datasets: CIFAR10, Reddit, IoT intrusion detection, MNIST, and FMNIST}, and show that it can effectively detect and eliminate malicious updates in FL without \mbox{deteriorating the benign performance of the global model.}

\end{abstract}
\vspace{-0.1cm}
\section{Introduction}
\label{sec:introduction}
\vspace{-0.2cm}
\noindent A machine learning framework is designed to learn from a single fused data collected from multiple data sources. This trainable data is comparable and homogeneous. However, in practice, data is heterogeneous and segregated across multiple decentralized devices. \changed{Learning a single machine learning model by using this scattered data is complex and challenging as it may disclose a user's identifiable and protected information. Federated Learning (FL) 
overcomes these drawbacks by enabling multiple distributed clients to learn a global model in a collaborative fashion~\cite{mcmahan2017,yang2018applied}. For instance, multiple hospitals can participate in training a global model for cancer classification without revealing individual patients' cancer records~\cite{li2020federated,sheller2018medical,xu2021federated}. Similarly, multiple smartphones could train together a word suggestion model without sharing the individually typed texts~\cite{mcmahan2017googleGboard}, or detect threats based on risk indicators~\cite{fereidooni2022fedcri}.}
In FL, each client locally trains a model on its private dataset and sends the parameters of this local model to a (global) server, which aggregates the different local models from the clients into a global model (see App.~\ref{app:fl} for more details). The server then responds by sending the aggregated model to each client in a single training round. By design, the global server is unaware of the training process being done locally on each client; thus, it is also \mbox{susceptible to poisoning attacks from malicious clients.}


\noindent\textbf{Poisoning Attacks and Defenses.} Previous works have shown that FL is prone to poisoning attacks as a malicious client (or clients) can inject malicious weights into the global server model during training~\cite{baehrens2010explain,bagdasaryan,blanchard17Krum,nguyen22Flame,shen16Auror,xie2020dba}. As a consequence, the performance of the global model on all or some subsets of predictive tasks becomes degenerated. In the so-called \emph{targeted} poisoning (or backdoor) attacks, the adversary's goal is to cause well-defined misbehavior of the global model on some trigger data points, i.e., predict a specific class if a particular pattern is present in the input data~\cite{bagdasaryan,nguyen2020diss,wang2020attack,xie2020dba}.\footnote{In contrast, non-targeted poisoning attacks aim to deteriorate the performance of the global model on all test data points~\cite{cao2020fltrust}.} Our focus in this paper is to mitigate such targeted backdoor attacks.

To detect/mitigate backdoor attacks, existing defenses leverage either the models' outputs (i.e., predictions on some validation data\footnote{As pointed out by Rieger~\etal, it is not realistic to assume validation data to be present on the aggregation server~\cite{rieger2022deepsight}.}), intermediary states (i.e., logits) of the models, and/or distance of the local models (i.e., \lnorm or cosine) with regard to the global model, or pairwise distances among the models.
However, current defenses have several shortcomings and are not sufficiently robust to defend against different classes of backdoor attacks. 
For instance, some defenses are bypassed when multiple different backdoors are simultaneously inserted by different malicious clients~\cite{shen16Auror}. Other defenses clip weights and add noise to negate the effect of malicious model updates, which reduces the benign performance of the global model~\cite{bagdasaryan,mcmahan2018iclrClipping,nguyen22Flame,sun2019can}, or they make specific assumptions such as (i) the adversary inserts malicious updates (backdoors) in each training round \cite{fung2020FoolsGold}, or (ii) the adversary attacks only at the end of the training~\cite{andreina2020baffle}, or (iii) the data of the benign clients having the same distribution~\cite{munoz19AFA,nguyen22Flame,yin2018byzantine}, or (iv) each benign client must \mbox{have a similar number of unique labels~\cite{rieger2022deepsight}.}  

\changed{Moreover, current state-of-the-art defenses against backdoor attacks make several assumptions about the underlying data and the adversary's adopted strategies, as well as they \emph{directly} employ client weights during detection. In this context, we encountered two main open challenges: First, how can we
compute an alternate, more generic, representation of client weights (or updates), such as a probabilistic measure, which will encompass all adjustments made to the updates due to any local training strategy (by the clients). Second, can we design an efficient detection/clustering algorithm that can leverage such a probabilistic measure to effectively filter out malicious updates in FL, without deteriorating the benign accuracy of the global model. We intuitively believe, and later empirically show, that designing a detection algorithm with such a generic probabilistic measure as one of its inputs provides several significant advantages over existing defense solutions. First, different local client training strategies will not affect the detection process at the global server. Consequently, the defense mechanism's detection phase will remain agnostic about an adversary's attack strategies. Second, utilizing distributions over client updates in the defense, instead of directly employing client weights, makes the detection process uninfluenced by the underlying local data distributions used for training. 
}


\noindent
\textbf{Our Goals and Contributions.}
To tackle the challenges outlined above, we present the design and implementation of \ourname, an unconventional and more general backdoor defense for FL that is based on a probabilistic machine learning framework. 
\changed{\ourname comprises of two main modules.} The first module computes a probabilistic measure of the client weights that is governed by the posterior of the \changed{\textit{Hierarchical Beta-Bernoulli}} process~\cite{thibaux2007hierarchical} (see~\sect\ref{sec:rationale}). The second module \changed{implements} a detection algorithm \changed{which employs this} probabilistic measure as an input to differentiate malicious and benign updates. The main idea is to utilize a probabilistic measure to determine the distribution of the incoming local client updates. Additionally, in each FL round, we compute the distribution of existing groups that were assigned client updates (clusters) or a new group (client updates can get assigned to a new group).
Then, we compute the (Jensen) divergence of these two distributions to detect malicious updates and compute the selected client's fit to an existing or a new cluster. The detection algorithm (described later) is mainly governed by the \textit{Chinese Restaurant Process (CRP)}, except that it uses \textit{\jensenDivergence} to compute clients' fit to the clusters.

\changed{The only work} in the literature that has employed similar Bayesian Non-Parametric (BNP) models in the context of FL is by Yurochkin et al.~\cite{yurochkin2019bayesian}, where BNP models, specifically the Beta-Bernoulli Process and the Indian Buffet Process, are used to reduce the communication overhead between the global server and the clients. They accomplished this by finding the common subset of neurons between the local clients selected in a training round and combining them to form a global model. In contrast to~\cite{yurochkin2019bayesian}, we use BNP models, specifically the Hierarchical Beta-Bernoulli process and CRP, for 
designing a \emph{defense} mechanism against backdoor attacks in FL. We stress that \changed{\cite{yurochkin2019bayesian} is vulnerable to backdoor attacks, as malicious training updates can easily be integrated into the global model.}

To the best of our knowledge, this is the first work that employs BNP modeling concepts to design an accurate and robust defense against backdoor attacks in FL. Our main contributions can be summarized as:

\begin{itemize}[leftmargin=*]
    \item We propose \ourname, a novel generic defense framework against backdoor attacks in FL that accurately and effectively detects backdoors without significantly impacting the benign performance of the aggregated model. \changed{Our proposed defense is relevant in many adversarial settings as, by design, the malicious update detection functionality utilizes distributions of client updates and, thus, is unaffected by any local client's strategy.
    }
    \item We take a new approach to the problem of mitigating backdoor attacks in FL by employing non-parametric Bayesian modeling in the design of the defense mechanism. \changed{To the best of our knowledge, existing defenses mainly consider the model updates as a set of vectors and matrices, and \emph{directly} administer these weights to filter out the malicious client updates~\cite{blanchard17Krum,fung2020FoolsGold,mcmahan2018iclrClipping,munoz19AFA,nguyen22Flame,shen16Auror}. Given the client weights, \ourname first estimates a probabilistic measure (such as the Beta posterior) that accurately captures the variations in the clients' weights and then uses a novel detection technique based on the Chinese Restaurant Process and Jensen-Divergence for identifying the poisoned models.}
    \item We extensively evaluate our framework on five benchmark datasets: CIFAR-10, Reddit, MNIST, FMNIST, and a real-world IoT network traffic dataset. \changed{We show that \ourname effectively mitigates different \sota as well as adaptive attacks, and accurately and effectively detects the backdoored models so that the benign performance of the aggregated model is not degraded, thus providing a significant advantage over \sota defenses.}
\end{itemize}

\vspace{-0.15cm}
\vspace{-0.1cm}
\section{Background and Intuition}
\label{sec:background}
\vspace{-0.2cm}

\noindent\changed{Our approach is modeled in two steps. First, to determine the probabilistic distributions of clients' updates, we make use of several statistical tools such as Beta Processes (BP), Hierarchical Beta Processes (HBP), and Bernoulli Processes (BeP).
Second, to design our detection algorithm, we outline an adaptation of the Chinese Restaurant Process (CRP), called CRP-Jensen, to detect and filter out malicious updates. Below, we briefly discuss the above two steps (see more technical details in the Appendix):
\noindent\\
\textbf{Determining probabilistic measure for client updates.} We compute the probabilistic measure for each client selected in an FL round to keep track of the adjustments made during each update. For this, we first draw a baseline probabilistic measure, denoted by the baseline Beta Process (BP), which is computed using the initial global model. Informally, a BP quantifies a subset of points (measure). We use BPs in this work to quantify the client updates and the global model by creating distributions over them.
    
A BP ($A$) is a stochastic process defined using two parameters: a concentration function $c$ over some space $\Omega = \mathbb{R}$
and a base measure $H$; denoted as $A \sim BP(c, H)$. In FL, the base measure $H$ can represent any distribution of the initial global model (see~\sect\ref{sec:rationale}), i.e., before the training starts, and a concentration function $c$ quantitatively characterizes the similarity between the input base measure ($H$) and the output random measure $A$ (because of the distribution over the random selection of elements in $\Omega$). In this work, $c$ determines the similarity between the input and the output distribution over $\Omega$, and $\Omega$ is a space of initial global model weights. The intuition here is to use this baseline BP, called baseline BP \emph{prior}, to form hierarchies of BP, called hierarchical BP prior, for $n$ different clients selected in the first FL round, \mbox{i.e., create $n$ sub-BP from the baseline BP. }

Informally, a \emph{prior} is the previous knowledge of an event before any new empirical data is observed and is typically expressed as a probability distribution or random measure, while a \emph{posterior} is the revised or updated knowledge of the event after considering the new data. Now, an HBP for each client $i$ is denoted as $A_{i} \sim BP(c_{i}, A$). In the subsequent iterations of the FL, these priors (as computed above) will be updated, based on the new client updates, to compute the so-called BP posteriors, i.e., update the $c_{i}$ and $H_{i}$ ($A_{i}$). In this work, we have assumed the new client updates in each round as the new data to update the previous knowledge of the BP priors, i.e., to compute the BP posteriors.
    
In this work, we flatten updates for each client $i$ to a one-dimensional vector having $l$ values, denoted as $W_{i}$. We assume that each value in this vector is drawn from a Bernoulli Process (BeP), given the client $i$'s BP random measure $A_{i}$. Informally, a BeP is a stochastic process with two possible outcomes: success or failure $-$ we use it in this work to show whether a client $i$'s update will have a particular value (or not), given its BP random measure $A_{i}$. In FL, each client updates its local model using the common aggregated global model sent by the global server. Hence, we postulate that each client update vector values are drawn from its corresponding BP random measure $A_{i}$, using BeP. Thus, a weight vector $W_{i}$ for client $i \in \{1,...,n\}$ is characterized by a Bernoulli Process, given as $W_{i} | A_{i} \sim \text{BeP}(A_{i})$. In other words, in $W_{i} = \{W_{i,1}, W_{i,2}, ..., W_{i,l}\}$, $l$ denotes the independent BeP draws over the likelihood function $A_{i}$.

Another reason to use BeP is that it has been shown in the literature that the Beta distribution is the conjugate of the Bernoulli distribution~\cite{broderick2018posteriors}. Hence, we do not have to use the computationally intensive Bayes' rule to compute the posteriors. We keep updating the corresponding HBP ($A_{i}$) for client $i$ using the conjugacy of the BP and the BeP, as given in \cite{thibaux2007hierarchical}. The posterior distribution of $A_{i}$ after observing $W_{i}$ is still a BP with modified parameters:
    \begin{equation}
        A_{i}|W_{i} \sim BP\left(c_{i}+l, \frac{c_{i}}{c_{i}+l} H + \frac{1}{c_{i} \cdot l} \sum_{l = 1}^{l}W_{i,l} \right) \label{eq_n:1}
    \end{equation}


\noindent\textbf{Designing the backdoor detection algorithm.} Next, we briefly describe how we adapt the Chinese Restaurant Process to detect malicious client updates. The CRP~\cite{socher2011spectral,blei2011distance,li2019tutorial} is an infinite (unknown number of clusters) mixture model in which customers (client's updates) are assigned tables (clusters) in a restaurant. In the context of FL, the clusters represent groups of incoming client updates. The customer can either sit at the already occupied tables (existing clusters) or at the new table (a new cluster is created). 
Our main idea, as discussed earlier, is to utilize a probabilistic measure to determine the distribution of the incoming local client $i$'s update. In addition, we also compute the distribution of the existing clusters of updates plus the new cluster. Then, we compute the \jensenDivergence between client $i$'s update distribution and each existing plus new cluster's distribution. Informally, \jensenDivergence (or Jensen-Shannon Divergence) is a measure of how similar two distributions are.
In consequence, we obtain a set of \jensenDivergence values. We take the maximum of this set to determine whether local client $i$ is malicious or not (intuition, as to why use maximum \jensenDivergence, is shown in \sect\ref{sec:rationale}). Based on this maximum \jensenDivergence value, we also determine the client $i$'s update cluster assignment. 
After the cluster is determined, we append the client $i$'s update to the selected cluster's list of client updates. Finally, we update the cluster's parameters, i.e., mean and standard deviation, using \textit{Chinese Restaurant Process (CRP)}.
This adaptation of the CRP is also referred to by us as CRP-Jensen. 

}

\vspace{-0.1cm}
\section{Adversary Model}
\label{sec:threat}
\vspace{-0.2cm}


\noindent \textbf{Attack Objectives.} The target system trains a Neural Network (NN) $f$ taking samples from a domain $\mathcal{D}$ 
as input and returning predictions from the set $\mathcal{L}$. The system realizes a function $f: \mathcal{D} \rightarrow \mathcal{L}$. 
The goal of the adversary \adversary is to inject a backdoor into the aggregated model making it predict a certain adversary-chosen label $l_\adversary \in \mathcal{L}$ for all samples that contain the backdoor trigger, called the \textit{trigger set} $\mathcal{D}_\adversary\subset\mathcal{D}$. The success of this objective is measured by calculating the accuracy for $\mathcal{D}_\adversary$.
The attack needs to be stealthy to prevent the backdoor from being detected. Therefore, \adversary needs to ensure that the attack does not affect the model's performance on the benign main task, i.e., changing the predictions of samples $d\in \mathcal{D}\setminus\mathcal{D}_\adversary$. For conducting such stealthy backdoor attacks, we assume that \adversary crafts poisoned model updates.  \adversary also needs to ensure that the poisoned model updates are indistinguishable from the benign model updates in terms of all the metrics that the aggregation server may use to detect poisoned models. As \adversary knows the defense mechanism deployed on the server side (see below), it suffices to make the poisoned model updates indistinguishable from the benign model updates in terms of the metrics that are used by the defense mechanism. \\

\vspace{-0.1cm}
\noindent \textbf{Attacker's Capabilities.} 
We assume \adversary to have the following capabilities to achieve its objectives:\\
\emph{1. Controlling malicious clients:} Aligned with existing work~\cite{andreina2020baffle,nguyen22Flame,shen16Auror}, we assume \adversary to fully control $\numberOfMaliciousClients < \frac{\numberOfClients}{2}$ clients where \textit{\numberOfClients} is the total number of participants. In particular, \adversary can arbitrarily manipulate the data and training process of the malicious clients. Therefore, besides poisoning the training data, \adversary can freely adapt the hyperparameters of the training process, and the loss function and can also scale the model updates before sending them to the aggregation server. \adversary does not control the benign clients. Moreover, it neither knows their training data
nor their model updates, although it can make a rough estimation of the benign model updates by training a model using the benign training data (i.e., without backdoors) of the malicious clients.\\
\emph{2. No control over the aggregation server.}  \adversary has complete knowledge of the global server's aggregation operations, including the deployed backdoor defenses. However, \adversary neither controls the server nor knows the parameters that are calculated by the server at runtime and can only interact with the server through the compromised clients. \changed{However, an adaptive \adversary can manipulate the model updates based on the knowledge of the deployed backdoor defense at the global server.}
\vspace{-0.1cm}
\section{Design}
\label{sec:rationale}
\vspace{-0.2cm}


\noindent \changed{In this section, we first discuss the requirements posed on \ourname due to the BNP nature of our defense}. Then, we outline the architecture of our \ourname defense mechanism and describe each component in detail.


\begin{figure}[t]
\centering
\includegraphics[width=\linewidth]{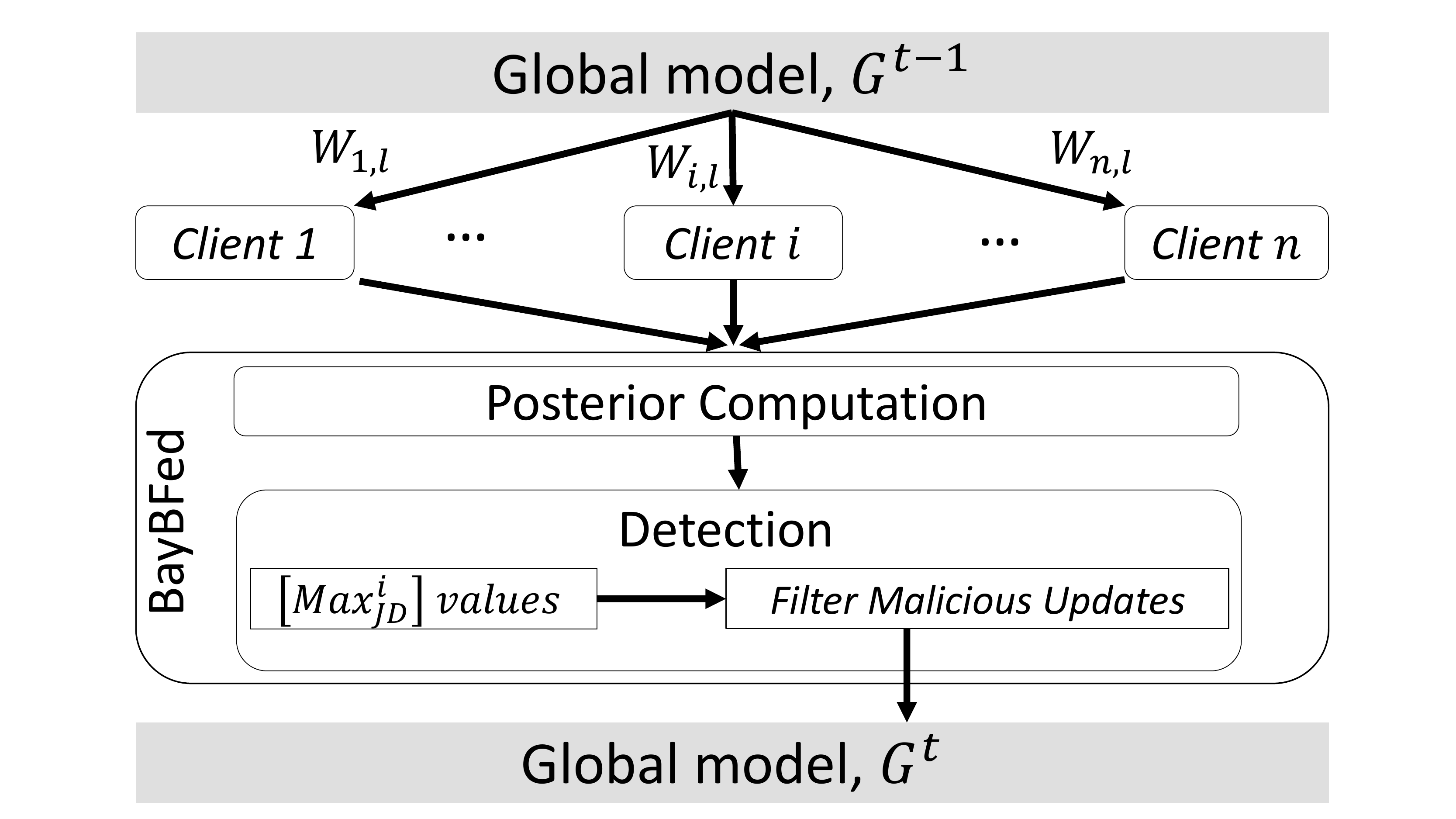}
\caption{High-level overview of \ourname.}
\label{fig:high_level}
\vspace{-0.2cm}
\end{figure}

\begin{figure*}[ht]
\centering
\includegraphics[width=0.9\textwidth]{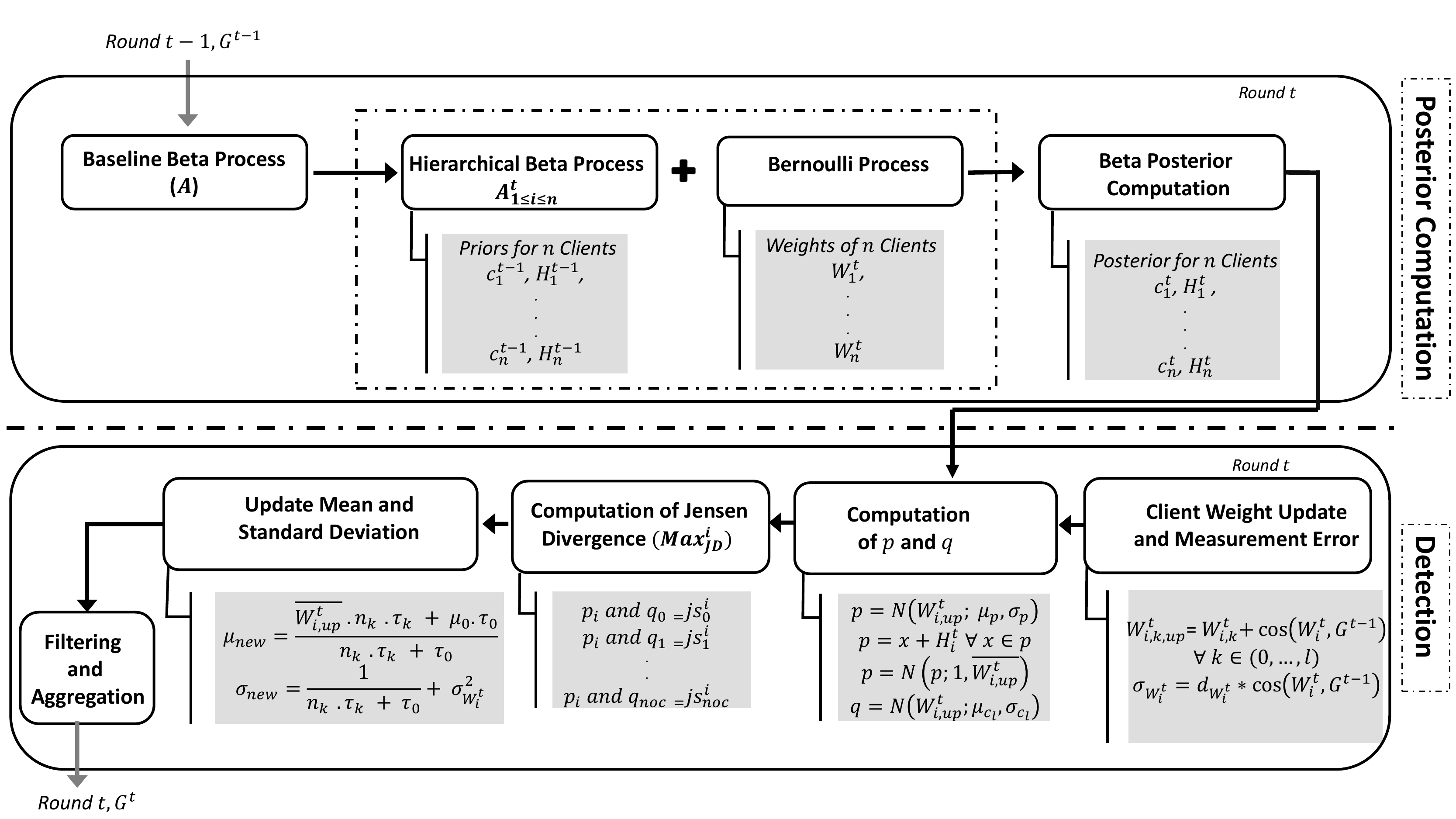}
\vspace{-0.1cm}
\caption{Illustration of \ourname's design, showing its two modules: Posterior computation and CRP-Jensen.}
\label{fig:algo}
\vspace{-0.35cm}
\end{figure*}
\vspace{-0.1cm}
\subsection{Requirements}\vspace{-0.2cm}
\label{subsec:require}
\noindent In BNP models, exchangeability (defined below) is a critical requirement that must be satisfied by a certain sequence of random variables to model different parameters such as priors and posteriors (see~\sect\ref{sec:background}). Since, the detection module (CRP-Jensen) takes client updates ($W_{i}$) as one of its inputs and its $l$ values are modeled by employing the Hierarchical Beta-Bernoulli Process (HBBP), both the client updates, $W_{i}$ and it's $l$ values should satisfy the exchangeability property.
Informally, the exchangeability property (of a sequence of random variables) states that the joint distribution of all the random variables remains the same for any permutation of random variables. 
Specifically, we identify the following two key requirements that we will use in the design of \ourname.

\noindent
\textbf{Requirement I. }\textit{For the posterior computation, a flattened client update vector is a sequence of random variables and should be drawn from an exchangeable set of choices.} 

\changed{We consider that the $l$ values in a client $i$'s update vector $W_{i}$ are drawn from an exchangeable set of choices.
The reason is, in Eq. \ref{eq_n:1}, we only utilize the summation of client $i$'s $l$ update values to update the base measure $H_{i}$. Hence, the order of the $l$ values in client $i$'s update will not affect the computation of $H_{i}$.}
Mathematically, a sequence of random variables $X_{1}, X_{2}, ..., X_{l}$ is called an exchangeable sequence, if the distribution of $X_{1}, X_{2}, ..., X_{l}$ is equal to the distribution of $X_{\pi_{1}}, X_{\pi_{2}}, ..., X_{\pi_{l}}$ for any permutation $(\pi_{1}, \pi_{2}, ..., \pi_{l})$. We consider $W_{i} = \{ W_{i, 1}, W_{i, 2}, ..., W_{i, l} \}$ to be an exchangeable sequence for the \mbox{computation of Beta posterior in \ourname.} 
\\

\noindent
\textbf{Requirement II. }\textit{For the detection algorithm employing CRP-Jensen, each client update is a sequence of random variables and should be drawn from an exchangeable set of choices.} 

CRP is an infinite mixture model which is used to assign data or samples to the mixtures (or clusters).
The data or samples are assumed to be drawn from an exchangeable set of choices. Hence, irrespective of the order in which the data arrives, their assignment to the mixtures or clusters (i.e., their seating arrangement in CRP) is not affected. 
\changed{In this work, we assign client $i$'s update $W_{i}$ to a cluster by employing CRP and \jensenDivergence (\jensenDivergenceShort). Thus, we consider $W_{i}$ to follow the exchangeability property. The reason is that client $i$'s local training does not depend on another client's local training. Thus, permuting the client updates $W_{i}$ or changing the order of the incoming client updates will not affect the output of the detection module.} Thus, in this work, we consider the incoming client updates $W_{i}$, where $1 \leq i \leq n$ and $n$ is the number of clients, as an exchangeable sequence.


\subsection{\ourname Components}
\label{subsec:modules_design}\vspace{-0.2cm}
\noindent In this section, we describe in detail the two main technical modules of \ourname, i.e., the posterior computation module and the detection module.
\subsubsection{Posterior Computation}
\label{subsubsec:posterior}

As briefly explained in \sect\ref{sec:background}, we compute Beta posteriors (using a concentration parameter and a base measure) to have a more generic representation of the client's weights, which can keep track of all the changes made in the client updates. 
The intuition here is to use the random measure parameters of the previous round $t-1$ (Beta prior), i.e., concentration parameter ($c^{t-1}$) and the base measure ($H^{t-1}$), and combine them with client updates ($W_{i}^{t}$) in round $t$, to compute the Beta posterior $c^{t}$ and $H^{t}$. This is done for each client $i$ selected in round $t$. Then, the updated base measure $H^{t}$ is utilized in the detection module to filter the poisoned updates. This process is repeated for the subsequent iterations of FL, until the model converges. A high-level overview of BayBFed's \mbox{architecture is depicted in Fig. \ref{fig:high_level}.}
Below, we outline a more detailed understanding of the components of the posterior computation as shown in Fig.~\ref{fig:algo}. \\

\noindent
\textbf{Baseline Beta Process (BP).} The first step is to create an initial or baseline BP ($A$) before any FL training starts. The goal is to use this baseline BP to create the sub-Beta priors using the HBP, for the clients selected in the first training round. 
\changed{In our experiments (as discussed later in \sect\ref{sec:results}), we initially choose a random baseline of $c=5$ and continuously update it based on the posteriors of client updates.} 
Further, we choose a base measure $H = \mathcal{N}(\mu_{p}, \sigma_{p})$ with $\mu_{p}$ equal to the mean of the flattened initial global model and $\sigma_{p}$ equal to the standard deviation of the flattened initial global deviation, i.e., populating $A$ with the initial global model weights. 
We assume that the data points (client updates) are normally distributed for the above mean and standard deviation computation.\\

\noindent
\textbf{Hierarchical Beta process (HBP).} The next step is to create hierarchies of the baseline BP for the clients selected in the first round. For a client $i$ selected in round $t$, HBP is used to define its BP as $A_{i}^{t} \sim BP(c_{i}^{t}, A)$.
In our experiments, before the training starts, we assign the same base measure of $H$ to each client selected in the first training round, and concentration parameters (for each client) are computed as random variables of a Poisson process
with parameter $c$. The Poisson process \cite{last2017lectures} creates randomized point patterns, and that is why we employ it to compute random concentration parameters ($c_{i}^{t}$) for each client $i$. After the first round, each client's concentration and base measure gets updated according to Eq.~\eqref{eq_n:1}.\\

\vspace{-0.1cm}
\noindent \textbf{Bernoulli Process (BeP). }In this work, BeP is defined as the draw of an exchangeable sequence of weights, $W_{i}^{t} = \{W_{i, 1}, W_{i, 2}, ..., W_{i, l}\}$, given the concentration parameter $c_{i}^{t}$ and the base measure $H_{i}^{t}$, i.e., Beta prior. This means the \mbox{$l$-dimensional} vector update of a client is considered to be the $l$ independent BeP. Given the client update $W_{i}^{t}$ at time $t$, we use Eq.~\eqref{eq_n:1}
to obtain \mbox{the Beta posterior of round $t$. }
The computed Beta posterior ($c_{i}^{t}$ and $H_{i}^{t}$) over client $i$'s update is integrated into the following detection module to determine whether incoming $W_{i}^{t}$ is malicious or not. 

\vspace{-0.1cm}
\subsubsection{Detection Module}
\label{subsubsec:filtering}
In this module, we design a variation of the CRP, called CRP-Jensen, to filter the poisoned updates sent by malicious clients (see \sect\ref{sec:background}).
CRP-Jensen ensures that all malicious updates are detected (and removed) without limiting the benign performance of the target global model.
The intuition here is to integrate the updated base measure ($H^{t}$) to compute the $p$ distribution of updated client weight, $W_{i,up}^{t}$, as shown in  Eq.~\eqref{eqn:2}.
Further, we compute a $q$ distribution across the updated client weight, $W_{i,up}^{t}$, for the existing clusters of the client updates or a new cluster (client update, $W_{i}^{t}$, can get assigned to a new cluster). Then, we compute a set of \jensenDivergenceShort between the client $i$'s $p$ and each cluster's $q$, obtaining a set (length: number of existing clusters + 1)  
of \jensenDivergenceShort values for each client $i$.

Next, we compute the maximum value ($Max_{JD}^{i}$) of this set and accordingly decide the cluster assignment for the corresponding clients. In the experiments (see~\sect\ref{sec:results}), we show that this value ($Max_{JD}^{i}$) varies significantly for malicious and benign updates. Thus, based on these acquired maximum \jensenDivergenceShort values, we filter out the malicious updates and perform the aggregation operation on the remaining benign client updates to obtain the global model $G^{t}$. Below, we outline a more detailed understanding of the \mbox{components of the detection module as shown in Fig.~\ref{fig:algo}.}\\

\noindent
\textbf{Client weight update and measurement error. }
First, we update each 
client's local model using the cosine angular distance ($cos(W_{i}^{t},G^{t-1})$) between the local model and the global model. The intuition for doing this is to integrate the effect of $cos(W_{i}^{t},G^{t-1})$ into the weights. The reason is that even though an adversary can manipulate the cosine angular distance, the poisoned weights have to differ (slightly) from the benign weights. Otherwise, the poisoned models will predict the correct label rather than the backdoor target label that \adversary chooses. Therefore, to run an effective attack, \adversary needs to simulate the weights in the backdoor direction. 
For the client's model $W_{i}^{t}$ with $l$ entries, where $W_{i,k,up}^{t}$ denotes the element at index k, the updated client weights ($W_{i,up}^{t}$) are computed as: 
\begin{equation}
    W_{i,k,up}^{t} = W_{i,k}^{t} + cos(W_{i}^{t},G^{t-1}) \quad \forall k \in \{0,\ldots,l\} \label{eqn:2}
\end{equation} 


In CRP, when a new sample is assigned to a cluster, the total error or variance is computed as a combination of two errors: the measurement error because of the new sample, and the errors due to already assigned samples. Thus, we compute the measurement error due to the new client's weight getting assigned to the specific cluster (in each round), given as:

\begin{equation}
    \sigma_{w_i^t} = d_{w_i^t} \cdot cos(W_{i}^{t},G^{t-1}) \label{eqn:3}
\end{equation}

\noindent where $d_{w_i^t}$ is the \lnorm between the global model in the previous round $G^{t-1}$ and client $i$ update in the current round $W_i^t$. Using the \lnorm as the measurement error has the same reasoning above for using the cosine angular distance. Even though an adversary can individually manipulate the \lnorm and cosine distance, there is still a correlation between the two that differs for the malicious and the benign weights. We found the above connections, shown in Eq.~\eqref{eqn:2} and Eq.~\eqref{eqn:3}, using our extensive experimental evaluations.
We integrate $W_{i,up}^{t}$ and $\sigma_{w_i^t}$ into the detection module to effectively eliminate all the malicious updates. \\

\noindent
\textbf{Computation of $p$ and $q$. }We then compute the two probability distributions $p$ and $q$, and use \jensenDivergenceShort to compute their similarity. Here, we integrate the current round base measure $H^{t}$ to compute $p$ distribution of client updates, such that the detection phase of the defense is not affected by any local client training strategy.
Thus, in round $t$, we compute each client's $p$ and each cluster's $q$ as given in Eq.~\eqref{eqn:4} and Eq.~\eqref{eqn:5}, respectively. 
\vspace{-0.2cm}
$$p = \mathcal{N}(W_{i, up}^{t}; \mu_{p}, \sigma_{p})$$
$$p = x + H^{t} \quad \forall x \in p$$
\begin{equation}
    p = \mathcal{N}(p; 1, \overline{W_{i, up}^{t}}) \label{eqn:4}
\end{equation}
\begin{equation}
\vspace{-0.2cm}
    q = \mathcal{N}(W_{i,up}^{t}; \mu_{c_{l}}, \sigma_{c_{l}}) \label{eqn:5}
\end{equation}

\noindent where, $\overline{W_{i, up}^{t}}$ is the mean of $W_{i,up}^{t}$ and $\mu_{p}$ and $\sigma_{p}$ are the mean and the variance of the initial global model, respectively. $H^{t}$ is the updated round $t$ base measure that is given as $\frac{c_{i}^{t-1}}{c_{i}^{t-1}+l} H^{t-1} + \frac{1}{c_{i}^{t-1} \cdot l} \sum_{i = 1}^{l}W_{i}^{t}$ (see Eq.~\eqref{eq_n:1}). $\mu_{c_{l}}$ and $\sigma_{c_{l}}$ are the clusters' mean and variance, respectively. Then for each client $i$, we compute the \jensenDivergenceShort of it's $p$ with each cluster's $q$.\\

\noindent \textbf{Computation of \jensenDivergence (\jensenDivergenceShort).} 
Next, we compute the \jensenDivergenceShort between each client's $p$ and each cluster's $q$. By computing the \jensenDivergenceShort of each client $i$'s $p$ values with each cluster's $q$ values, we get a set of: $\{(p_{i}, q_{0}):js_{0}^{i}, (p_{i}, q_{1}):js_{1}^{i}, ..., (p_{i}, q_{noc}):js_{noc}^{i}\}$. Then, we compute $Max_{JD}^{i}$: max$(js_{0}^{i}, js_{1}^{i}, ..., js_{noc}^{i})$ to output the assigned cluster of client $i$ weights $W_{i, up}^{t}$ and to decide whether it's a malicious update or a benign update. Here, $noc$ is the total number of clusters formed yet. \\

\noindent
\textbf{Mean and standard deviation update. }
In the previous step, we computed the client's assigned cluster. Now, we  update the mean and the variance of that particular cluster according to the equations:

\begin{equation}
    \mu_{new} = \frac{\overline{W_{i, up}^{t}} n_{k} \tau_{k}+ \mu_{0} \tau_{0}}{n_{k} \tau_{k}+ \tau_{0}} \label{eqn:6}\vspace{-0.1cm}
\end{equation}

\begin{equation}
    \sigma_{new} = \frac{1}{n_{k}\tau_{k}+ \tau_{0}} + \sigma_{w_i^t}^{2}  \label{eqn:7}
\end{equation}

\noindent where, $n_{k}$ is the number of client updates already assigned to it, $\tau_{k}$ represents the precision of the cluster, $\mu_{0}$ and $\tau_{0}$ represent the initial mean and the precision assumed for the new clusters. $\sigma_{w_i^t}^{2}$ is the variance or the measurement error introduced by the new addition of the client update and is computed according to Eq.~\eqref{eqn:3}. 
\begin{figure}[tb]
\center
\includegraphics[width=0.85\linewidth]{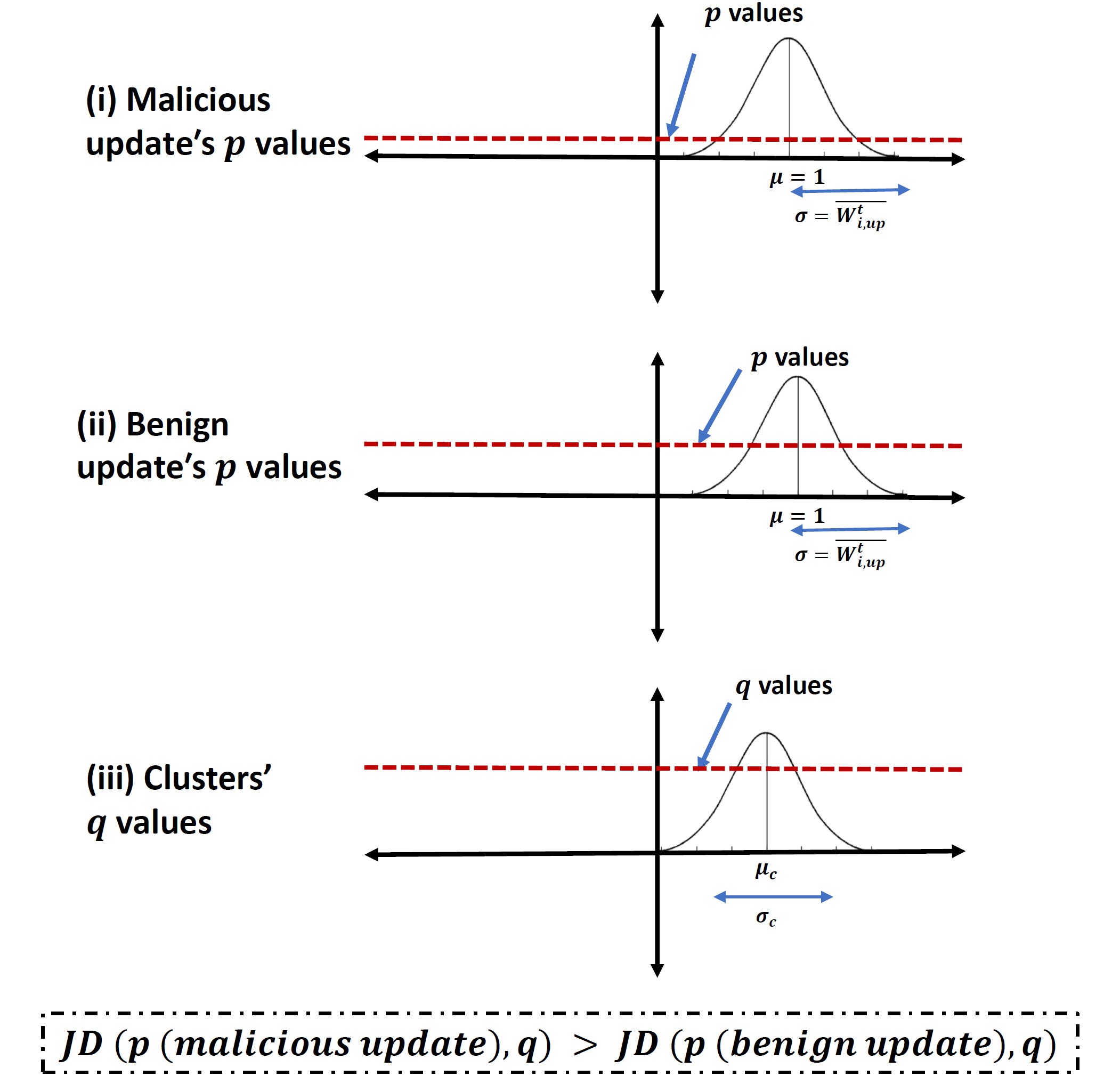}
\caption{$p$ and $q$ distribution value range for (i) the malicious updates, (ii) the benign updates, and (iii) the clusters.}
\label{fig:JSD21}
\end{figure}\\

\noindent
\textbf{Filtering and Aggregation. }Finally, we examine the patterns of the malicious updates based on the computed $Max_{JD}^{i}$, which differs significantly from the benign updates. 
\changed{We encountered two patterns of malicious updates $Max_{JD}^{i}$ in our experiments, as discussed later in \sect\ref{sec:results}: (i) the $Max^i_{JD}$ computed for malicious updates are much greater than that for the benign updates, and (ii) the computed $Max^i_{JD}$ values for the malicious updates are similar to each other. For pattern (i), we observed that the $Max_{JD}^{i}$ value for benign updates is less than the average of all the clients' $Max_{JD}^{i}$ values (experimentally evaluated). Therefore, conditioned on this observation, we filtered the malicious updates during the detection phase of \ourname. For pattern (ii), we check if $Max_{JD}^{i}$ of the incoming update is already present in the set of computed $Max_{JD}^{i}$ for the $n$ clients; if yes, we do not include the concerned malicious client's update in the final aggregation of \mbox{updates to output the global model $G^{t}$.}} \\

\changed{\noindent\textbf{Intuition for the filtering step. } To understand the relationship between the computed maximum \jensenDivergenceShort values $Max^i_{JD}$ (as outlined above) and the benign/malicious nature of the updates, we conduct experiments utilizing a diverse set of datasets (see \sect\ref{sec:setup}). In these experiments, we illustrated the $Max^i_{JD}$ values for malicious and benign updates and observed that $Max^i_{JD}$ values of malicious and benign updates differ significantly. Fig.~\ref{fig:JSD21} gives an intuition of why $Max^i_{JD}$ differs for the malicious and benign updates by examining the range of $p$ distribution values for the client updates against the clusters' $q$ distribution values. In Fig.~\ref{fig:JSD21}, the (i) plot demonstrates the malicious update's $p$ values spanning area, the (ii) plot demonstrates the benign update's $p$ values spanning area, and the (iii) plot demonstrates the clusters' $q$ values spanning area, as observed from the experiments conducted. As seen in these plots, the benign update's $p$ values lie at a larger distance than the malicious update's $p$ values (as the values in $p$ are either equal to or very close to zero). Thus, the distance between the $p$ values of malicious updates, and the  $q$ values of the clusters is greater than the distance between the  $p$ and $q$ values of benign updates. In other words, \jensenDivergenceShort($p$ (malicious update), $q$) $>$ 
\jensenDivergenceShort($p$ (benign update), $q$). Hence, maximum \jensenDivergenceShort is used as a metric to identify malicious updates \mbox{and assign them to the new cluster.}}

\subsection{\ourname WorkFlow and Algorithms}
\label{subsec:algorithms}
\vspace{-0.2cm}

\changed{
\noindent \ourname's workflow and detection algorithm have been outlined in Algorithms~\ref{algo:jensen} and \ref{algo:detection}. Here, $\mu_{0}$ is the assumed initial mean of the clusters and  $\sigma_{0}^{2}$ is the assumed variance corresponding to mean $\mu_{0}$. $\sigma_{w_i^t}^{2}$ is the measurement error of the client update $W_{i}^{t}$ and is computed as shown in Eq.~\eqref{eqn:3}. Thus, total measurement error or the variance is computed as shown in Eq.~\eqref{eqn:7}. If a new cluster is formed, it will have a normal distribution with mean $\mu_{0}$ and the combined variance of $\sigma_{0}^{2}$ + $\sigma_{w_i^t}^{2}$. The set $\{\mu_{c_{l}}^{t}$, $\sigma_{c_{l}}^{t}\}$ represents the mean and standard deviations of the $c_{l}$ clusters at time t.
We start Algorithm \ref{algo:jensen} by looping through the number of rounds of FL training as shown in line $2$. In each round, we initialize an empty array, $Max_{JD}^{stored}=[]$, to store the $Max_{JD}^{i}$ values of the clients. $n$ clients are selected for the training, and we loop through each client $i$ (line $4$) to determine its cluster. We then compute $W_{i, up}^{t}$ and $\sigma_{w_i^t}$ according to equations \eqref{eqn:2} and \eqref{eqn:3}, respectively (lines $7$). If $noc=0$, then it's the first round, and the first client is assigned to the first cluster (line $9$), and accordingly, this new cluster's $\mu_{new}$ and $\sigma_{new}$ are updated (line $10$). If $noc \neq 0$, then for each existing cluster plus the new one (line $12$), we do the following: first, we compute $p$ and $q$ (line $13$), second, we compute the \jensenDivergenceShort of each client's $p$ and each cluster's $q$ (line $14$), third, we compute the maximum of the obtained \jensenDivergenceShort set ($Max_{JD}^{i}$) and decide the assigned cluster (line $15$) according to this value, and finally append it to the array $Max_{JD}^{stored}$ (line $16$). Either the client will be assigned to one of the already formed clusters (line $17$) or it will be assigned to a new cluster (lines $20, 21$). After each FL round, Algorithm \ref{algo:detection} ($DetectFilter()$) is called which takes input $Max_{JD}^{stored}$ (line $25$) and returns the filtered client updates, \textbf{\textit{fcp}}. $FedAVG(\textbf{\textit{fcp}})$ (defined in Appendix~\ref{app:fl}) algorithm is then performed to aggregate the filtered client updates and finally update the global model. 
}

\begin{algorithm}[tb]
\caption{\ourname's workflow.}
\label{algo:jensen}

\scalebox{0.9}{
\begin{minipage}{1.12\columnwidth}
\begin{algorithmic}[1]
\STATE \textbf{Input:} $\mu_{0}$, $\sigma_{0}^{2}$, $\sigma_{w_i^t}^{2}$, $\tau_{0} = \frac{1}{\sigma_{0}^{2}}$, $\tau_{w}=\frac{1}{\sigma_{w_i^t}^{2}}$, $noc$. 
\FOR{ each round till the model converges }
    \STATE Initialize an array to store $Max_{JD}^{i}$ in each round, $Max_{JD}^{stored}=[]$.
    \FOR{ \textit{i} 1 $\leftarrow$ \textbf{to} n }
        \STATE Draw any client update $W_{i}^{t}$.
        \STATE Compute $d_{w_i^t}$ and $cos(W_{i}^{t},G^{t-1})$ . 
        \STATE Update $\sigma_{w_i^t}^{2}$ $\leftarrow$ $d_{w_i^t} \cdot cos(W_{t,i},G^{t-1})$ and compute $W_{i,up}^{t}$.
        \IF{$noc$ == 0}
        \STATE Assign $\textit{c}_{0} \leftarrow W_{i}^{t}$. 
        \STATE Update $\mu_{new}$ and $\sigma_{new}$ with, $n_{k}=1$. 
        \ENDIF
        \FOR{ \textit{$c_{l}$} $\leftarrow$ \textbf{to} \textit{noc+1}}
        \STATE Compute $p$ and $q$. 
        \STATE Compute the \jensenDivergenceShort by $p$ and $q$ and store the values.
        \STATE Decide the cluster, $c_{i}$ according to $Max_{JD}^{i}$.
        \STATE Append $Max_{JD}^{i}$ to $Max_{JD}^{stored}$.
        \IF{ $\textit{c}_{l,i}=c_{l}$}
        \STATE Update $W_{i}^{t}$ assigned cluster, $\{\mu_{c_{l}}^{t}$, $\sigma_{c_{l}}^{t}\}$ according to  $\textit{c}_{l,i}=c_{l}$.
        \ELSE 
        \STATE Increment: $noc=noc+1$, a new cluster is formed.
        \STATE Set $\textit{c}_{l,i}=noc$ and assign $W_{i}^{t}$ to it. Append this new cluster to the vector of non-empty clusters.
        \ENDIF
        \ENDFOR
    \ENDFOR
    \STATE Call $DetectFilter()$, \textbf{\textit{fcp}} = $DetectFilter(Max_{JD}^{stored})$. 
    \STATE Perform $FedAVG(\textbf{\textit{fcp}})$ and update the global model.
\ENDFOR
\end{algorithmic}
\end{minipage}}
\end{algorithm}

\begin{algorithm}[tb]
\caption{Detection and Filter Algorithm, $DetectFilter$.}
\label{algo:detection}

\scalebox{0.9}{
\begin{minipage}{1.12\columnwidth}
\begin{algorithmic}[1]
\STATE \textbf{Input: $Max_{JD}^{stored}$ containing $Max_{JD}^{i}$ values, total clients, $n$}
\STATE \textbf{Output: Filtered client updates}
    \STATE Initialize an array to store filtered client updates in each round, $W_{filtered}=[ ]$.
    \STATE Compute $Max_{JD}^{avg}$ = sum($Max_{JD}^{stored}$)/$n$
    \FOR{$ \mbox{i}=1 \mbox{ to } n$}
        \IF{$Max_{JD}^{i}$ not in $Max_{JD}^{stored}$ }
            \STATE $W_{filtered}$.append($W_{i}^{t}$)
        \ENDIF
        \IF{ $Max_{JD}^{i} < Max_{JD}^{avg}$}
            \STATE $W_{filtered}$.append($W_{i}^{t}$)
        \ENDIF
    \ENDFOR
\STATE \textbf{Return $W_{filtered}$} 
\end{algorithmic}
\end{minipage}}
\end{algorithm}

\vspace{-0.1cm}
\section{Experimental Setup}
\label{sec:setup}
\vspace{-0.2cm}
\noindent We employ the machine learning framework PyTorch to conduct our experiments and use the existing defenses~\cite{blanchard17Krum,fung2020FoolsGold,shen16Auror,munoz19AFA,yin2018byzantine,mcmahan2017learning,nguyen22Flame} as baseline models to comparatively analyze the performance of \ourname. \changed{Aligned with previous work on backdoor attacks~\cite{andreina2020baffle,bagdasaryan,nguyen22Flame}, we use the attacks provided by Bagdasaryan \etal~\cite{bagdasaryan} and Wang \etal~\cite{wang2020attack} to implement the \ConstrainAndScale and Edge-Case backdoor attacks.} Below, we provide the configurations of the different datasets and the accuracy and precision metrics we use to evaluate the performance of \ourname.  

\noindent
\textbf{Datasets. }\changed{To show the generality of our results and the representative nature of \ourname across models/data from different domains, we evaluate the proposed defense mechanism by designing two attacks (see Tab.~\ref{tab:success_attacks}) on three popular FL applications: (i) image classification, (ii) word prediction, and (iii) IoT network intrusion detection. To facilitate an equitable comparison of \ourname with state-of-the-art backdoor attack approaches~\cite{bagdasaryan,nguyen22Flame}, we align the datasets, setups, and NN architectures employed in our comparative evaluation \mbox{with the ones used by these research efforts.}}\\
\emph{Image Classification (IC):} \changed{We use the popular benchmark datasets MNIST, FMNIST, and CIFAR-10 in our experiments. As these datasets are frequently used for evaluating FL and backdoor attacks and defenses~\cite{bagdasaryan,cao2021provably,fung2020FoolsGold,guerraoui2018hidden,khazbak2020mlguard,li2020learning,mcmahan2017,munoz19AFA,nguyen22Flame,rieger2022deepsight,wang2020attack,wu2020mitigating,xie2021crfl,fereidooni2021safelearn,rieger2022close}, it enables us to perform an equitable comparative analysis of our approach with other state-of-the-art approaches in the literature.} All three consist of samples belonging to one out of ten classes, handwritten digits in the case of MNIST, articles of clothing in the case of FMNIST, and objects (airplanes, cars, birds, etc.) in the case of CIFAR-10. The CIFAR-10 dataset consists of 50K training and 10K test images, while MNIST and FMNIST datasets each consist of 60K training and 10K test images. As the NN architecture, a light-weight version of Resnet-18 is used for CIFAR-10~\cite{bagdasaryan}, a simple CNN is used for MNIST~\cite{cao2021provably}, and a three-layer fully connected NN with \textit{relu} activations is used for FMNIST.\\
\emph{Word Prediction (WP):} To evaluate \ourname for a complex Natural Language Processing (NLP) application such as word prediction, we use the Reddit dataset consisting of all posts from November 2017. \changed{Aligned with the work of Bagdasaryan~\etal, we considered each author's posts as a local dataset and only the 50K most frequent words. A Long Short-term Memory (LSTM) model is used to predict the next word~\cite{bagdasaryan}.}\\
\changed{\emph{Network Intrusion Detection (NIDS):} Further, we evaluate \ourname for the FL-based NIDS \diot~\cite{nguyen2019diot} system using four real-world network traffic datasets, kindly shared with us by Nguyen~\etal~\cite{nguyen2019diot,nguyen22Flame} and Sivanathan~\etal~\cite{sivanathan2018UNSWdata}. The datasets consist of network traffic of multiple smart home and office settings. Aligned with previous work~\cite{nguyen22Flame,rieger2022deepsight}, we converted the network packets into symbols based on their features, such as source and destination ports, protocols, and flags. To simulate a distributed FL setting, we split the dataset into 100 local datasets, each consisting of symbols between 2K and 3K, which were extracted from the network packets. The NN is trained to predict the next probabilities for each possible symbol (network packet). The NN consists of 2 Gated-Recurrent-Unit layers followed by a fully connected linear layer, as defined by Nguyen~\etal~\cite{nguyen2019diot}. }

\noindent
\textbf{Evaluation metrics. }We compute four metrics to estimate the accuracy and precision of \ourname.\\
\textit{True Positive Rate (TPR): }This metric specifies how accurately the defense is able to detect the poisoned model updates. The total number of correctly identified poisoned updates are called True Positives ($TP$) and the number of poisoned model updates discerned as benign model updates are called False Negatives ($FN$). Thus, $TPR = \frac{TP}{TP+FN}$.\\ 
\textit{True Negative Rate (TNR): }This metric determines how accurately the defense is able to detect the benign model updates. The total number of correctly identified benign model updates are called True Negatives ($TN$) and the number of benign updates discerned as poisoned updates are called False Positives ($FP$). Thus, $TNR = \frac{TN}{TN+FP}$.\\  
\textit{Backdoor Accuracy (BA): }This metric is used to measure the accuracy of the model on the triggered inputs. Specifically, it measures the fraction of triggered samples where the model predicts the adversary's chosen label.\\
\textit{Main Task Accuracy (MA): }This metric is used to measure the accuracy of the model on its benign main task. It represents the fraction of benign inputs for which the model provides correct predictions. 

\section{Experimental Results}
\label{sec:results}
\vspace{-0.2cm}
\noindent Next, we empirically illustrate the effectiveness of \ourname against two \sota attacks~\cite{bagdasaryan,wang2020attack} and compare its efficacy against various \sota defense mechanisms. Further, we show how $Max_{JD}^{i}$ varies for the malicious and benign model updates. Finally, we demonstrate the robustness of \ourname for various adversarial attack parameters and sophisticated backdoor injection strategies.

\vspace{-0.15cm}
\subsection{Overall Performance}
\label{subsec:performace}
\vspace{-0.25cm}

\begin{table}
\caption{Backdoor Accuracy ($BA$) and Main Task Accuracy ($MA$) of \ourname compared to two \sota attacks. All values are represented as percentages.}
\label{tab:success_attacks}
\vspace{-0.1cm}
\centering
\scaleTable{
\begin{tabular}{|c|r|r|r|r|r|r|r}
\hline
\textbf{Attacks} & \textbf{Dataset} & \multicolumn{2}{|c|}{\textbf{No Defense}} & \multicolumn{2}{|c|}{\textbf{\ourname}} \\ \hline
  & & \multicolumn{1}{c|}{BA} & \multicolumn{1}{c|}{MA} & \multicolumn{1}{c|}{\hspace{0.45cm}BA \hspace{0.45cm}}& \multicolumn{1}{c|}{MA} \\ \hline
\multirow{3}{*}{\ConstrainAndScale~\cite{bagdasaryan}}   & Reddit & 100.0 & 22.6 & 0.0 & 22.6 \\ \cline{2-6}
                              & CIFAR-10 & 100.0 & 90.5 & 0.0 & 92.2 \\ \cline{2-6}
                              & \changed{MNIST} & \changed{43.0}  & \changed{96.5}  & \changed{0.0}  & \changed{96.0}\\\cline{2-6}
                              & \changed{FMNIST} & \changed{71.0}  & \changed{85.5}  & \changed{2.0}  & \changed{85.3}\\\cline{2-6}
                              & IoT-Traffic & 100.0 & 100.0 & 0.0 & 100.0 \\ \hline
 Edge-Case~\cite{wang2020attack}                    & CIFAR-10 &  33.16 & 88.42& 4.02 & 82.82 \\ \hline
\end{tabular}}
\end{table}

\noindent
\textbf{Attack Strategies. }\changed{The effectiveness of \ourname against two \sota model poisoning attacks, the \ConstrainAndScale~\cite{bagdasaryan} and the Edge-Case backdoor~\cite{wang2020attack} is shown in Tab.~\ref{tab:success_attacks}.
As we have assumed that an adversary can fully control the malicious clients (and thus the code on the clients), he is not restricted or constrained in terms of the employed attack strategy. In addition to attacks during training, our adversary can also adopt a runtime strategy to make the attack more stealthy. 
}

As can be seen in Tab. 2, \ourname functions optimally against \ConstrainAndScale attacks by filtering out all poisoned updates (BA = 0\%). At the same time, the $MA$ remains approximately equal to the benign setting $MA$. \changed{It should be noted that if the MA is less than 100\%, misclassifications of the model can be counted in favor of the backdoor, especially if the model wrongly predicts the backdoor target. As already pointed out by Rieger~\etal~\cite{rieger2022deepsight}, this phenomenon primarily occurs for image scenarios with pixel-based triggers. It causes the BA to be slightly higher than 0\% for backdoor-free models. In the case of an Edge-Case attack, the BA before the attack and after \ourname integration is 11.22\% and 4.02\%, respectively. However, without defense, the BA achieves 33.16\%.}
\begin{figure*}[htbp]
     \centering
	\begin{subfigure}[b]{0.2\textwidth}
         \centering
         \includegraphics[width=\textwidth, trim=.2cm .5cm 0.2cm 0]{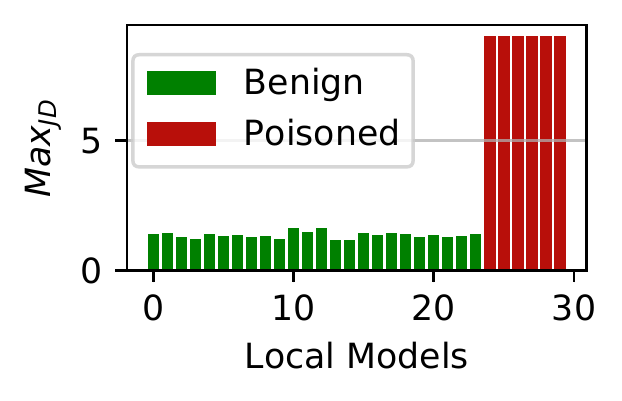}
         \caption{\nonIid = 0.0}
  	    \label{fig:cifar_iid_0}
    \end{subfigure}
    \begin{subfigure}[b]{0.2\textwidth}
         \centering
         \includegraphics[width=\textwidth, trim=0.2cm 0.5cm 0.2cm 0]{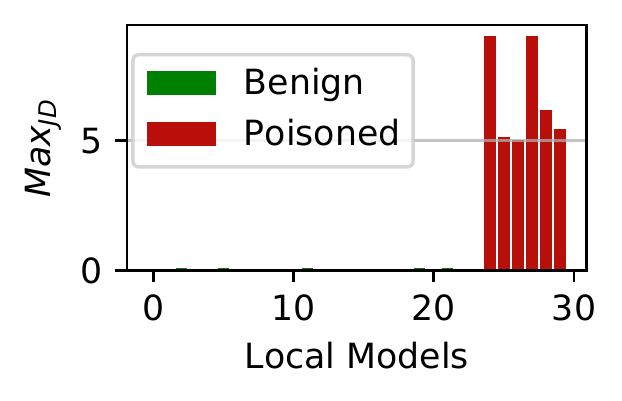}
         \caption{\nonIid = 0.5}
         \label{fig:cifar_iid_5}
    \end{subfigure}
    \begin{subfigure}[b]{0.2\textwidth}
         \centering
         \includegraphics[width=\textwidth, trim=0.2cm 0.5cm 0.2cm 0]{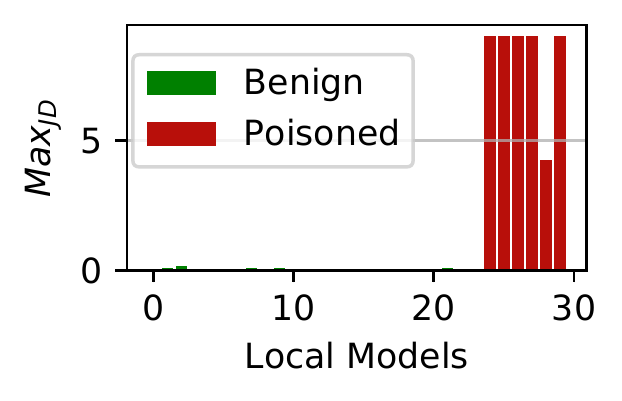}
         \caption{\nonIid = 1.0}
         \label{fig:cifar_iid_10}
    \end{subfigure}\vspace{-0.15cm}
    \caption{Effect of different \nonIid rates on the maximum \jensenDivergence ($Max_{JD}^{i}$) for CIFAR-10 dataset.}
        \label{fig:cifar_iid}\vspace{-0.3cm} 
\end{figure*}

\begin{figure*}[htbp]
     \centering
	\begin{subfigure}[b]{0.2\textwidth}
         \centering
         \includegraphics[width=\textwidth, trim=0.4cm 0.5cm 0.4cm 0]{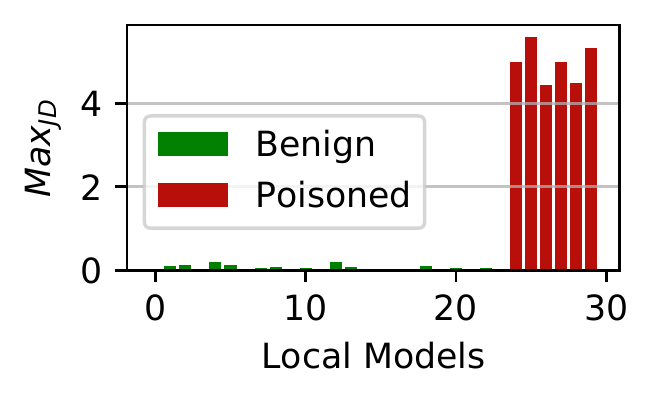}
         \caption{PMR = $20\%$}
  	\label{fig:cifar_pmr_6}
     \end{subfigure}
      \begin{subfigure}[b]{0.2\textwidth}
         \centering
         \includegraphics[width=\textwidth, trim=0.4cm 0.5cm 0.4cm 0]{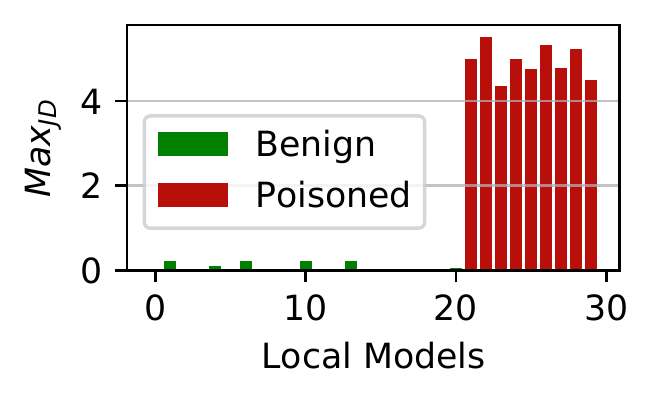}
         \caption{PMR = $30\%$}
         \label{fig:cifar_pmr_9}
     \end{subfigure}
      \begin{subfigure}[b]{0.2\textwidth}
         \centering
         \includegraphics[width=\textwidth, trim=0.4cm 0.5cm 0.4cm 0]{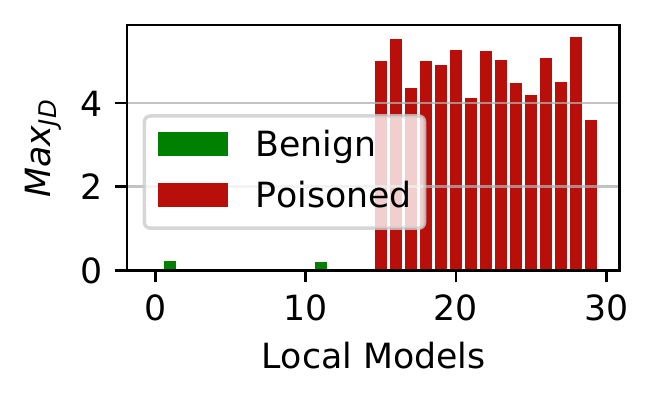}
         \caption{PMR = $50\%$}
         \label{fig:cifar_pmr_15}
     \end{subfigure}\vspace{-0.15cm}
        \caption{Effect of different PMR rates on the maximum \jensenDivergence ($Max_{JD}^{i}$) for the CIFAR dataset.}
        \label{fig:cifar_pmr}\vspace{-0.4cm}
\end{figure*}

\noindent
\textbf{Baseline Models. }\changed{We compare \ourname against seven \sota defense mechanisms present in the literature: Krum~\cite{blanchard17Krum}, FoolsGold~\cite{fung2020FoolsGold}, Auror~\cite{shen16Auror}, AFA \cite{munoz19AFA}, DP~\cite{yin2018byzantine}, Median~\cite{mcmahan2017learning} and FLAME~\cite{nguyen22Flame}.} We implement the \ConstrainAndScale attack against all the defenses and compare the output statistics in terms of the $BA$ and $MA$. As illustrated in Tab.~\ref{tab:success_defenses}, \ourname outperforms all these defense mechanisms. These results show that the existing defense mechanisms either lack the precision in removing all the poisoned updates or limit the $MA$ of the global model. \changed{Further, these defense mechanisms perform accurately when specific assumptions about the data and attack scenarios are satisfied. For instance, in the case of Krum~\cite{blanchard17Krum}, which selects a single model as an aggregated model, a poisoned model is chosen when an attacker circumvents Krum. Therefore, the aggregated model is entirely replaced by a poisoned model, achieving 100\% BA. Similarly, another defense FoolsGold~\cite{fung2020FoolsGold}, is effective for the highly \nonIid Reddit dataset but fails when their clients have similar data.} 
It should be noted that \ourname achieved a TPR and TNR of 100\% in all three scenarios.

\noindent\textbf{Impact on the MA.}
\changed{For the IC application CIFAR-10 dataset, we observe that the \ConstrainAndScale attack lowers the MA from 92.6\% (FedAVG without attack) to 90.5\% (FedAVG). Krum, FoolsGold, Auror, DP, and Median techniques achieve a $MA$ of 56.7\%, 52.3\%, 26.1\%, 78.9\%, and 50.1\%, respectively, which is considerably lower than the benign setting $MA$. In contrast, \ourname has a $MA$ of 92.2\%, which shows that it works significantly better for IC applications.} For the WP application, Krum and DP have a decreased $MA$ of 9.6\% and 18.9\%, compared to the highest $MA$ of 22.6\%. In this case as well, \ourname performs much better and achieves a $MA$ of 22.6\%. For the IoT-Traffic dataset, every defense has decreased $MA$. In this case, even small drops in MA need to be avoided due to the nature of this application. The reason is due to the high number of network packets in this scenario; even a small number of false alerts will annoy the user, causing them to ignore the alerts. For example, the defense technique FLAME results in a drop of 0.2\%, which causes 2 out of every 1000 packets to be misclassified. As a result, when a high amount of network packets is sent, the user will receive a high number of alerts. It should be noted \ourname recognizes all benign and malicious models correctly ($TPR=100\%$ and $TNR=100\%$) in all three scenarios, thus, comparatively performing better than the other defense mechanisms such as FLAME. For example, FLAME excludes benign models; in the NIDS scenario, FLAME wrongly excludes 17 benign models, which \mbox{might be problematic in the case of highly \nonIid data.}\\
\noindent\textbf{Backdoor updates removal. }Krum and Auror fail to remove poisoned updates in all three applications, as these defenses exhibit a $BA$ of 100\%. FoolsGold eliminates all the poisoned updates in the Reddit dataset ($BA=0.0\%$). However, it fails to remove them in the CIFAR-10 and IoT-Traffic datasets, as it achieves a $BA$ of 100\% in those cases. For the AFA defense, it works accurately for CIFAR-10 ($BA=0.0\%$) but is ineffective for the Reddit ($BA=0.0\%$) and IoT-Traffic ($BA=0.0\%$) datasets. In contrast, \ourname significantly outperforms these defenses as it can remove poisoned updates ($BA=0.0\%$) for all the datasets.
\begin{table}
\caption{Backdoor Accuracy ($BA$) and Main Task Accuracy ($MA$) of \ourname compared to \sota defenses for the  \ConstrainAndScale attack. All values are represented as percentages.}
\label{tab:success_defenses}
\vspace{0.15cm}
\centering
\scaleTable{
    \begin{tabular}{|l|r|r|r|r|r|r|}
    \hline
    \multirow{2}{*}{\textbf{Defenses}} & \multicolumn{2}{|c|}{\textbf{Reddit}} & \multicolumn{2}{|c|}{\textbf{CIFAR-10}} & \multicolumn{2}{|c|}{\textbf{IoT-Traffic}} \\ \cline{2-7}
                                & \multicolumn{1}{c|}{BA} & \multicolumn{1}{c|}{MA} & \multicolumn{1}{c|}{BA} & \multicolumn{1}{c|}{MA} & \multicolumn{1}{c|}{BA} & \multicolumn{1}{c|}{MA}  \\ \hline
    Benign Setting     & \multicolumn{1}{c|}{-}  & 22.6 & \multicolumn{1}{c|}{-} & 92.6 & \multicolumn{1}{c|}{-} & 100.0 \\ \hline
    No Defense         & 100.0 & 22.6 & 100.0 & 90.5 & 100.0 & 100.0 \\ \hline 
    Krum~\cite{blanchard17Krum}                 & 100.0 & 9.6 & 100.0 & 56.7 & 100.0 & 84.0 \\ \hline
    FoolsGold~\cite{fung2020FoolsGold}        & 0.0 & 22.5 & 100.0 & 52.3 & 100.0 & 99.2 \\ \hline
    Auror~\cite{shen16Auror}                    & 100.0 & 22.5 & 100.0 & 26.1 & 100.0 & 96.6 \\ \hline
    AFA~\cite{munoz19AFA}                       & 100.0 & 22.4 & 0.0 & 91.7 & 100.0 & 87.4\\ \hline
    DP~\cite{bagdasaryan}                       & 14.0 & 18.9 & 0.0 & 78.9 & 14.8 & 82.3 \\ \hline
    Median~\cite{yin2018byzantine}              & 0.0 & 22.0 & 0.0 & 50.1 & 0.0 & 87.7\\ \hline
    FLAME~\cite{nguyen22Flame}                  & 0.0 & 22.3 & 0.0 & 91.9 & 0.0 & 99.8\\ \hline
    \textbf{\ourname}                & \textbf{0.0} & \textbf{22.6} & \textbf{0.0} & \textbf{92.2} & \textbf{0.0} & \textbf{100.0} \\ \hline 
    \end{tabular}}
    \vspace{0.1cm}
\end{table}

Next, we discuss the impact of two critical experimental parameters on each of the considered applications and datasets in this paper: poisoned model rate ($PMR$) and degree of \nonIid data.
$PMR$ represents the fraction of \numberOfMaliciousClients malicious clients per total clients \numberOfClients. Thus, $PMR=\frac{\numberOfMaliciousClients}{\numberOfClients}$.
\nonIid represents the percentage of \nonIid data at each client. A \nonIid value of 0 means that the data is independently and identically distributed, \nonIid = $1.0$ implies that the data of different clients differ significantly and are distinguishable. For the IC application, we simulate experiments for both \nonIid degrees and $PMR$ (see~\sect\ref{subsec:ic}). However, for the Reddit and the IoT datasets, changing the \nonIid degree is not meaningful since this type of data has a natural distribution, as every client obtains data from different Reddit users or traffic chunks from different IoT devices. Thus, we only simulate experiments for different $PMR$ for these two datasets. We will also show the impact of these two parameters on $Max_{JD}^{i}$ for each client (see~\sect\ref{sec:rationale}) and prove that it differs significantly for the benign and poisoned updates.
\vspace{-0.1cm}
\subsection{\ourname Statistics for CIFAR-10}
\label{subsec:ic}\vspace{-0.2cm}
\noindent In this section, we evaluate the impact of \nonIid rate and $PMR$ on the CIFAR-10 dataset. First, we demonstrate the trend of $Max_{JD}^{i}$ for both the malicious and benign clients with respect to each of these parameters. Then, we illustrate the impact of \nonIid rate and $PMR$ on \ourname's performance by quantifying different metrics as stated in \sect\ref{sec:setup} and also compare it against the no defense scenario. 

\noindent
\textbf{Illustration of $Max_{JD}^{i}$. }
The impact of the degree of \nonIid data and $PMR$ on $Max_{JD}^{i}$ for each client is shown in Fig.~\ref{fig:cifar_iid} and Fig.~\ref{fig:cifar_pmr}, respectively. We select a total of 30 (\numberOfClients = 30) clients for both the \nonIid and $PMR$ experimental analysis. For \nonIid analysis, we test \nonIid $\in \{0.0, 0.5, 1.0\}$ and set $PMR=0.2$. Thus, the number of malicious clients equals 6 (\numberOfMaliciousClients = 6). For $PMR$ analysis, we test $PMR \in \{0.2, 0.3, 0.5\}$, i.e., when $\numberOfMaliciousClients$ equals 6, 9, and 15 and set \nonIid = 0.7. As illustrated in Fig.~\ref{fig:cifar_iid} and Fig.~\ref{fig:cifar_pmr}, the $Max_{JD}^{i}$ value for benign clients differs significantly from that of malicious clients. Hence, \ourname easily filters out all the malicious client updates, achieving a $BA$ of zero while keeping the $MA$ of the global model intact.

\begin{figure}[b]
     \centering

      \begin{subfigure}[b]{0.2\textwidth}
         \centering
         \includegraphics[width=\textwidth, trim=0.3cm 0.4cm 0.3cm 0]{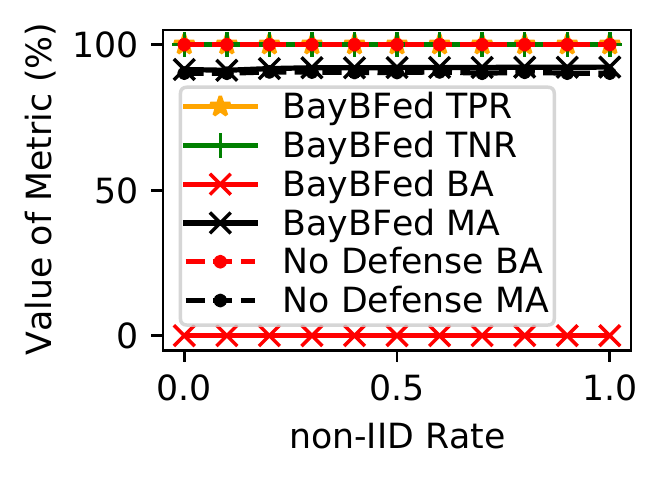}
         \caption{\nonIid Rate}
         \label{fig:cifar_iid_impact}
     \end{subfigure}
     \begin{subfigure}[b]{0.2\textwidth}
         \centering
         \includegraphics[width=\textwidth, trim=0.3cm 0.4cm 0.3cm 0]{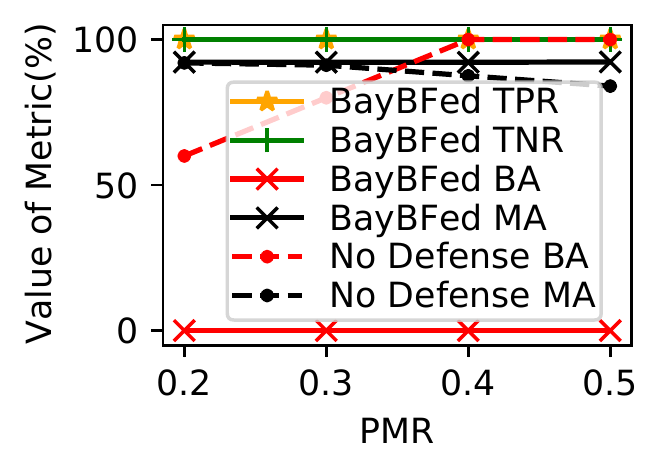}
         \caption{PMR}
         \label{fig:cifar_pmr_impact}
     \end{subfigure}\vspace{-0.15cm}
        \caption{Impact of the poisoned model rate $PMR =\frac{\numberOfMaliciousClients}{\numberOfClients}$ and \nonIid rate on \ourname for the IC application.}
        \label{fig:cifarMetrics}
\end{figure}

\noindent
\textbf{Effect of the degree of \nonIid Data. }
To study the impact of \nonIid data on \ourname, we conduct experiments for the \ConstrainAndScale attack on the CIFAR-10 dataset. \changed{Following recent work~\cite{wang2020attack,fang2020local,rieger2022deepsight,nguyen22Flame}, we prepare the \nonIid data by varying the number of images assigned to a particular class for each client. Precisely, we form 10 groups corresponding to the ten classes of CIFAR-10. Then, clients in each group are allocated a fixed fraction of images, depending on the \nonIid degree of that group's label, while allocating the remaining images to each client randomly. Mainly, for \nonIid = $0.0$, the samples of all clients followed the same distribution and were chosen randomly from all classes. However, for \nonIid = $1.0$, the samples of each client were only chosen from the samples belonging to the main class of this client.}
Fig.~\ref{fig:cifar_iid_impact} compares the impact of the degree of \nonIid data in terms of $BA$ and $MA$ for the plain FedAVG without defense (No Defense $BA$, No Defense $MA$) and the impact on \ourname ($BA$, $MA$). 
Fig.~\ref{fig:cifar_iid_impact} also shows the computed $TPR$ and $TNR$ for \ourname in this setting. As one can observe, we obtain $TPR=100\%$, indicating \ourname achieved $BA$ = 0, i.e., all the poisoned models were detected and filtered out before the aggregation. In addition, \ourname achieved $TNR=100\%$, indicating it correctly identified all the benign updates, thus getting \mbox{approximately $MA=92.2\%$ for all the \nonIid rates.} \\
\textbf{Effect of different $PMR$ rates. }
Fig.~\ref{fig:cifar_pmr_impact} shows the impact of different $PMR$ rates on \ourname. We consider $PMR$s of 0.2, 0.3, 0.4, and 0.5. Hence, \numberOfMaliciousClients equals 6, 9, 12, and 15. We use the same metrics that we used for \nonIid rates to evaluate \ourname against different $PMR$s. In this experiment, we achieve results similar to the ones we obtained for different \nonIid rates. This demonstrates that \ourname is efficient and accurate in eliminating all the poisoned updates for different data distributions while keeping the benign accuracy of the model intact.

\begin{figure*}[htbp]
    \centering
	\begin{subfigure}[b]{0.2\textwidth}
         \centering
         \includegraphics[width=\textwidth, trim=0.2cm 0.4cm 0.2cm 0]{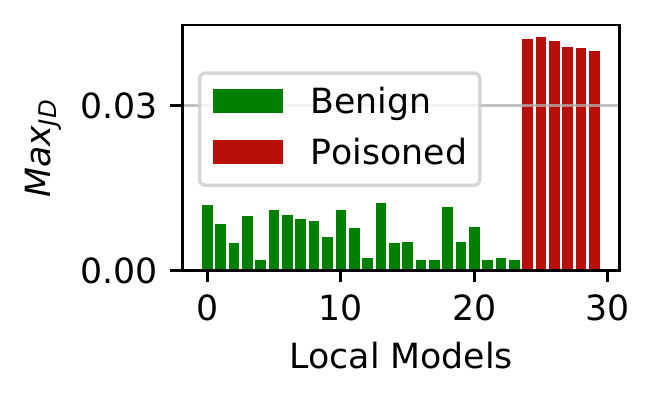}
         \caption{PMR$=20\%$}
  	\label{fig:WP_pmr_6}
     \end{subfigure}
    \begin{subfigure}[b]{0.2\textwidth}
         \centering
         \includegraphics[width=\textwidth, trim=0.2cm 0.4cm 0.2cm 0]{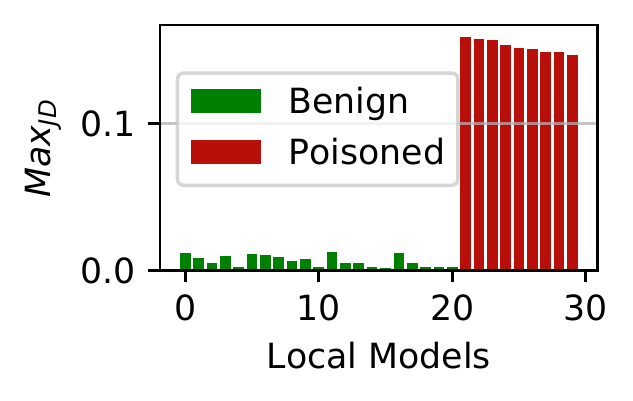}
         \caption{PMR$=30\%$}
         \label{fig:WP_pmr_9}
    \end{subfigure}
    \begin{subfigure}[b]{0.2\textwidth}
         \centering
         \includegraphics[width=\textwidth, trim=0.2cm 0.4cm 0.2cm 0]{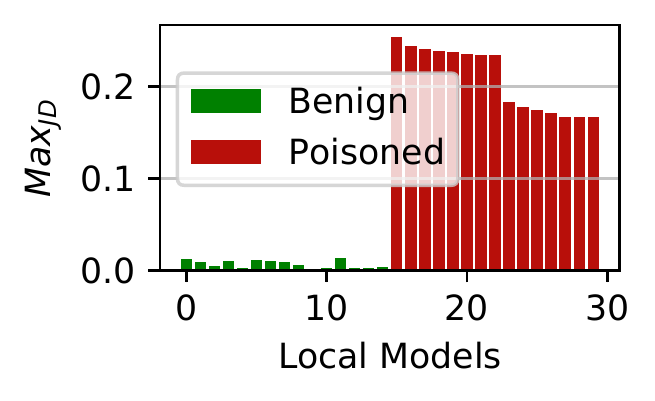}
         \caption{PMR$=50\%$}
         \label{fig:WP_pmr_15}
     \end{subfigure}
     \vspace{-0.2cm}\caption{Effect of different PMR rates on the maximum \jensenDivergence ($Max_{JD}^{i}$) for the Reddit dataset.} 
        \label{fig:WP_pmr}\vspace{-0.2cm}
\end{figure*}

\begin{figure*}[htbp]
    \centering
	\begin{subfigure}[b]{0.2\textwidth}
         \centering
         \includegraphics[width=\textwidth, trim=0.2cm 0.4cm 0.2cm 0]{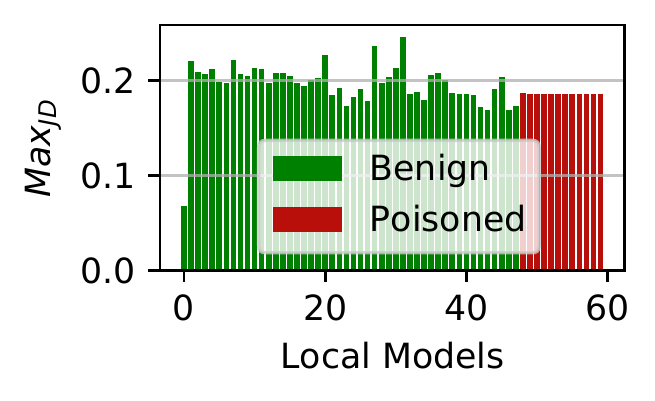}
         \caption{PMR = $20\%$}
  	\label{fig:iot_pmr_12}
     \end{subfigure}
      \begin{subfigure}[b]{0.2\textwidth}
         \centering
         \includegraphics[width=\textwidth, trim=0.2cm 0.4cm 0.2cm 0]{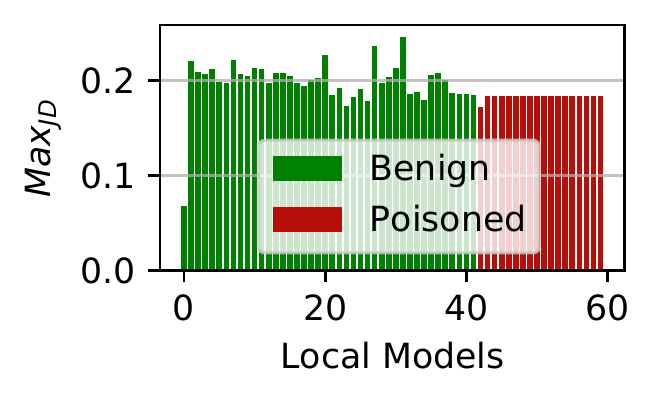}
          \caption{PMR = $30\%$}
         \label{fig:iot_pmr_18}
     \end{subfigure}
      \begin{subfigure}[b]{0.2\textwidth}
         \centering
         \includegraphics[width=\textwidth, trim=0.2cm 0.4cm 0.2cm 0]{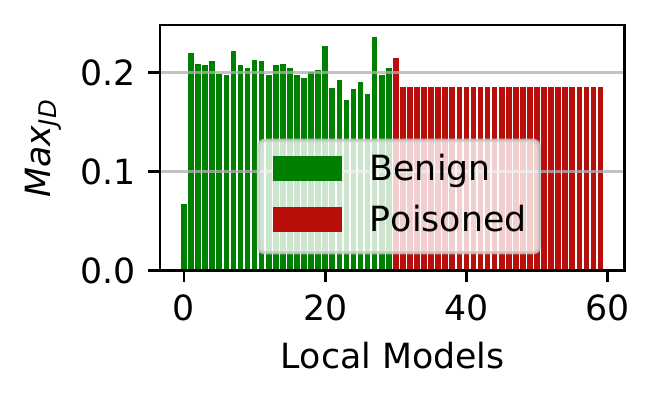}
          \caption{PMR = $50\%$}
         \label{fig:iot_pmr_30}
     \end{subfigure}
        \vspace{-0.2cm}\caption{Effect of different PMR rates on the maximum \jensenDivergence ($Max_{JD}^{i}$) for the IoT-Traffic dataset.}
        \label{fig:iot_pmr}\vspace{-0.4cm}
\end{figure*}


\vspace{-0.1cm}
\subsection{\ourname Statistics for WP}
\label{subsec:wp}\vspace{-0.2cm}

\noindent This section evaluates the impact of $PMR$ on the Word Prediction application. First, we demonstrate the trend of $Max_{JD}^{i}$ for both the malicious clients and the benign clients. Then, we illustrate the impact of $PMR$ on \ourname's performance by quantifying different metrics as stated in \sect\ref{sec:setup} and also compare it against the no defense scenario.

\noindent
\textbf{Illustration of $Max_{JD}^{i}$. }
In this setting, we also select 30 clients who can participate in each training round, 
and demonstrate the impact of varying $PMR$ values ($0.2, 0.3, 0.5$) on the clients' $Max_{JD}^{i}$. As outlined in Fig.~\ref{fig:WP_pmr}, the $Max_{JD}^{i}$ values of malicious and benign client updates differ significantly. Thus, \ourname accurately identified all the poisoned updates, achieving a $BA = 0.0\%$ and $MA=22.6\%$. \\
\textbf{Effect of different PMR rates. }
Next, we evaluate the effectiveness of \ourname, compared against no defense $BA$ and $MA$, for different $PMR$ values ($0.2, 0.3, 0.4, 0.5$). The results of this experiment are shown in Fig.~\ref{fig:wp_pmr_impact}. These results indicate that \ourname obtained a $TPR=100\%$ and a $TNR=100\%$, for all $PMR$ values. Moreover, it successfully identified all the poisoned and benign updates for different $PMR$ values and achieved a $BA=0\%$ and the highest possible $MA$ of benign setting, i.e., $MA = 22.6\%$.

\vspace{-0.1cm}
\subsection{\ourname Statistics for NIDS}
\label{subsec:iot}
\vspace{-0.2cm}
\noindent This section evaluates the impact of different $PMR$ rates on the NIDS application. Here, we randomly select 60 clients who can participate in each training round. \changed{It should be noted that since NIDS models have a lesser number of parameters, training time is reduced. Thus, we evaluated more clients than WP and IC models/applications. However, we set the same $PMR$ in all scenarios. Hence, it did not impact the experimental results (except for the experiments where we considered different $PMR$s).}
The number of benign and malicious clients varies based on the selected $PMR$ value, specifically, for $PMR$ values 0.2, 0.3, 0.4, and 0.5, \numberOfMaliciousClients is 12, 18, 24, and 30, respectively. First, we demonstrate the trend of $Max_{JD}^{i}$ for both the malicious and benign clients and then illustrate the impact of $PMR$ on \ourname compared to the no defense scenario.

\begin{figure}[tb]
     \centering
      \begin{subfigure}[b]{0.2\textwidth}
         \centering
         \includegraphics[width=\textwidth, trim=0.2cm 0.4cm 0.2cm 0]{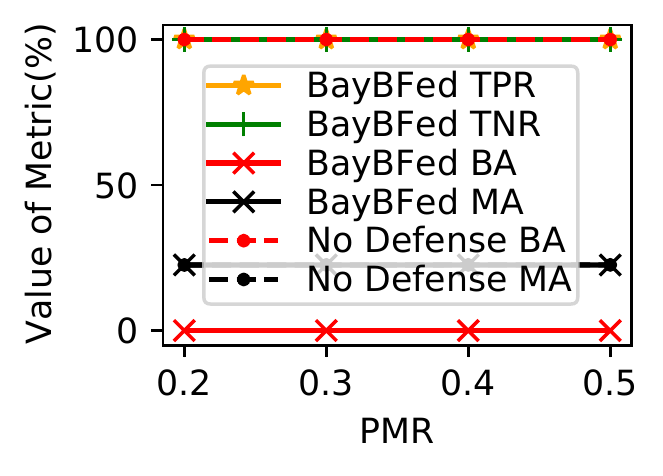}
         \caption{WP}
         \label{fig:wp_pmr_impact}
     \end{subfigure}
     \begin{subfigure}[b]{0.2\textwidth}
         \centering
         \includegraphics[width=\textwidth, trim=0.2cm 0.4cm 0.2cm 0]{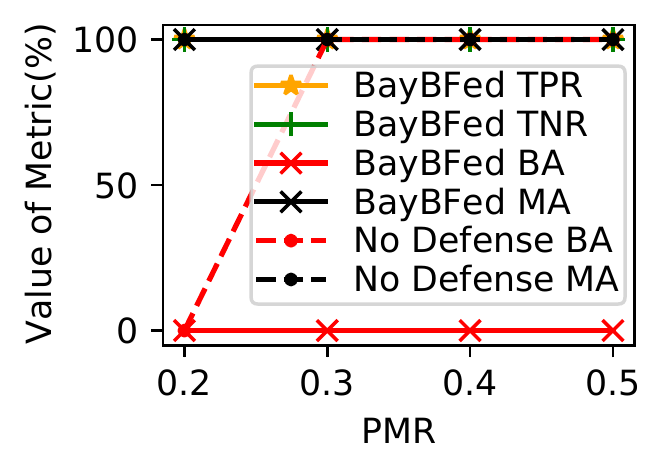}
         \caption{NIDS}
         \label{fig:iot_pmr_impact}
     \end{subfigure}\vspace{-0.2cm}
        \caption{Impact of the poisoned model rate $PMR =\frac{\numberOfMaliciousClients}{\numberOfClients}$ on the evaluation metrics.}
        \label{fig:pmr_impact}
\end{figure}

\noindent
\textbf{Illustration of $Max_{JD}^{i}$. }
Fig.~\ref{fig:iot_pmr} illustrates the impact of different $PMR$ values on $Max_{JD}^{i}$ for each client. This plot illustrates the sequence of $Max_{JD}^{i}$ for the poisoned updates, and one can observe that they are equal and different from benign updates. By employing this pattern of the $Max_{JD}^{i}$, \ourname was accurately able to filter out all the poisoned updates, thus, attaining a $BA$ of $0\%$. \\
\textbf{Effect of different PMR rates. }
Next, we compute the $TPR$, $TNR$, $BA$, and $MA$ metrics to evaluate the effectiveness of \ourname compared against the no defense $BA$ and $MA$, for different $PMR$ values.
Results for this set of experiments are shown in Fig.~\ref{fig:iot_pmr_impact}. By using the computed maximum \jensenDivergence values for each client, \ourname is able to achieve $TPR=100\%$, $TNR=100\%$, $BA =0\%$, and $MA=100\%$. Hence, \ourname performs optimally \mbox{for the NIDS application as well.}

\changed{\subsection{\ourname Statistics for FMNIST and MNIST}
\label{subsec:fmnist}
\vspace{-0.2cm}
\noindent Further, we evaluate the impact of different \nonIid and $PMR$ rates on the FMNIST and MNIST datasets.
We used the same setup that we used for the CIFAR-10 dataset (see~\sect\ref{subsec:ic}). 
In all the experiments with FMNIST and MNIST, $Max_{JD}^{i}$ values of malicious and benign client updates differ significantly, as observed for CIFAR-10. Thus, \ourname accurately identified all the poisoned updates, achieving a $BA$ of $0\%$.  For detailed FMNIST and MNIST results, please refer to App.~\ref{app:fmnist} and App.~\ref{app:mnist}, respectively.}

\vspace{-0.1cm}
\subsection{Effect of Other Factors on \ourname}
\label{subsec:strategies}
\vspace{-0.2cm}
\noindent Next, we conduct additional experiments with \ourname by varying four other parameters: (i) number of clients (hence, the number of malicious clients), (ii) backdoor injection strategies, (iii) poisoned data rates ($PDR$), and (iv) client order. \changed{Additionally, we also assess the trade-off between model accuracy and defense evasion for an adaptive attacker.} $PDR$ represents the fraction of injected poisoned data in the overall poisoned training dataset. Our goal in conducting these experiments is to show that \ourname is robust against these factors in detecting backdoor attacks in FL.

\begin{figure}[t]
    \centering\vspace{-0.03cm}
    \begin{subfigure}[b]{0.48\textwidth}
         \centering
         \includegraphics[width=\textwidth,trim=0 0.6cm 0 0]{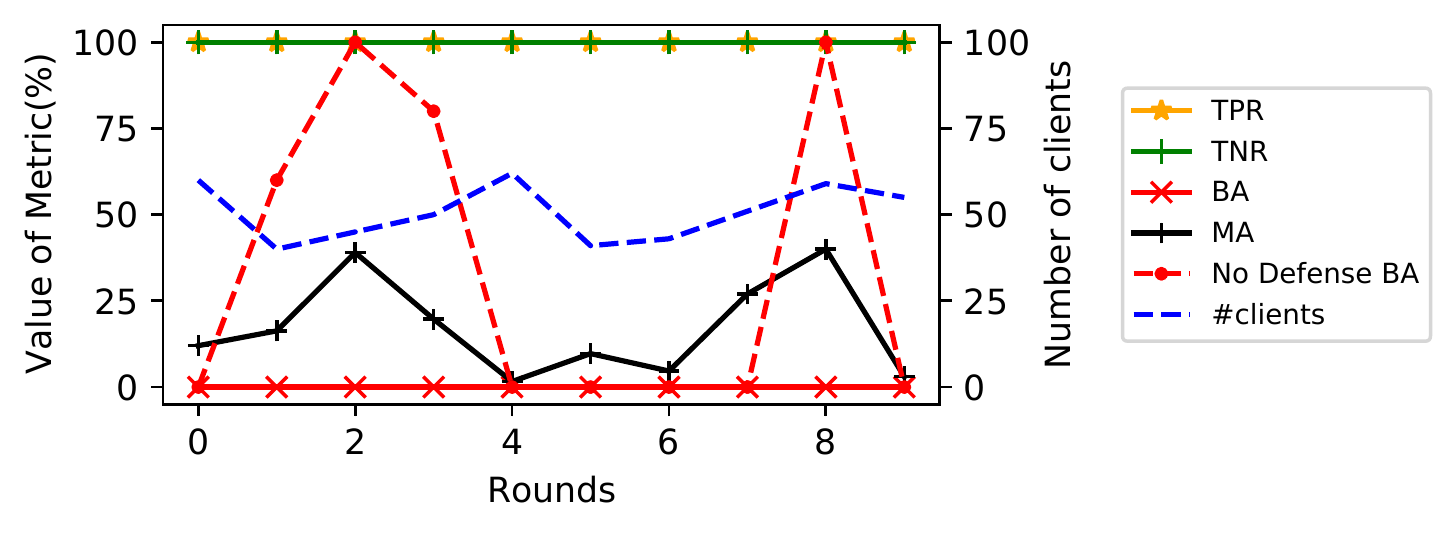}
         \vspace{-0.4cm}
         \caption{IC}
  	    \label{fig:bd_nodefense_ic}
    \end{subfigure}
    \begin{subfigure}[b]{0.48\textwidth}
         \centering
         \includegraphics[width=\textwidth,trim=0 0.6cm 0 0]{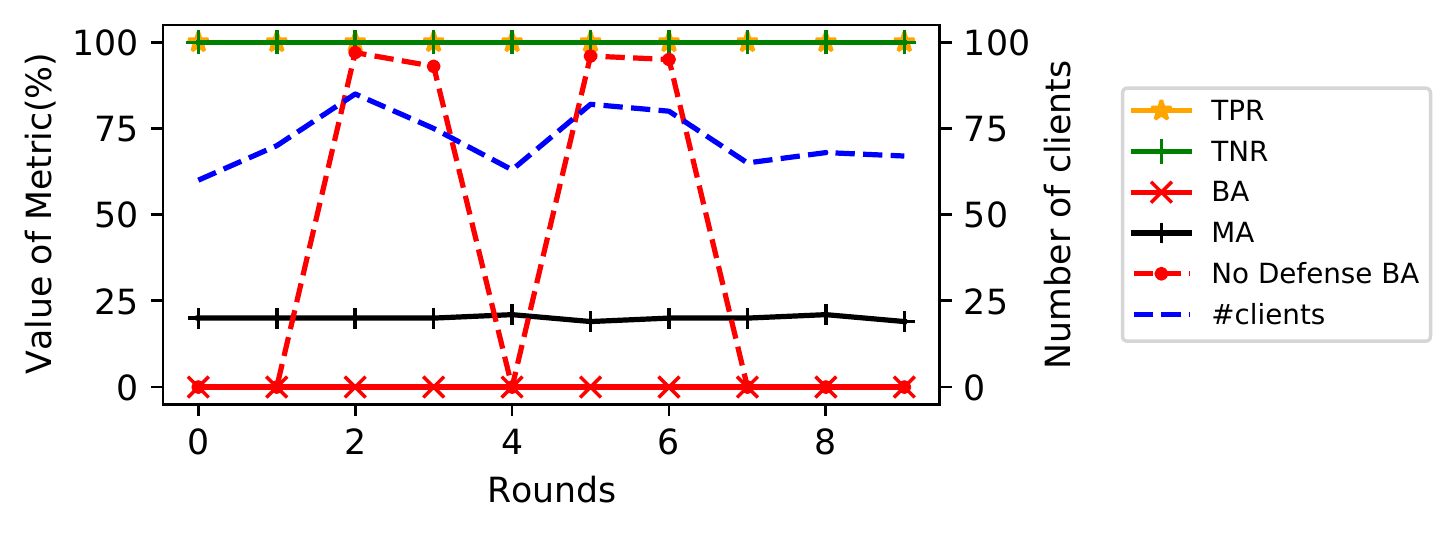}
         \vspace{-0.3cm}
         \caption{NIDS}
         \label{fig:bd_nodefense_iot}
    \end{subfigure}
  \caption{Impact of the number of clients on \ourname vs No Defense for different datasets.}
    \label{fig:bd_nodefense}
\end{figure}

    \noindent\textbf{Number of clients.}
    In this experiment, we evaluate the impact on the performance of \ourname by varying the number of clients, thus, the $PMR$. The results are outlined in Fig.~\ref{fig:bd_nodefense}. In each round, we select a random number of clients ranging from $40$ to $90$. 
    We conduct this experiment for the IC (Fig.~\ref{fig:bd_nodefense_ic}) and NIDS (\ref{fig:bd_nodefense_iot}) applications. In both cases, \ourname achieved a $BA$ of $0\%$, thereby showing that it is effective in eliminating all the backdoors compared to the no defense scenario.

    \noindent\textbf{Different injection strategies. }
    An adversary (\adversary) can inject multiple backdoors at the same time in order to make the backdoor more difficult to detect, thus making the poisoned models harder to distinguish from benign ones in \nonIid scenarios. \changed{We perform four experiments for the NIDS application, where each client is trained to inject 1 to 4 backdoors. Existing work \cite{rieger2022deepsight} has shown that the attack efficiency significantly reduces as the number of backdoors increases, and we observed the same pattern during our experiments. 
    Hence, four backdoors were considered a good number (of backdoors) that provided reasonable attack efficiency. Our evaluations show that \ourname was able to defend against and mitigate all the introduced backdoors effectively, thus achieving a 0\% BA.}
    
    \noindent\textbf{Different Poisoned Data Rates ($PDR$).}
    In this experiment, we consider an adversary that is capable of poisoning the data to launch backdoor attacks. We evaluate this attack on the CIFAR-10 and IoT-Traffic dataset for three different values of $PDR$: 0.05, 0.1, and 0.5, i.e., 5\%, 10\%, and 50\% of the training dataset is poisoned. For the CIFAR-10 dataset, we set $n = 30$ and $PMR = 0.2$, and for the IoT-Traffic dataset, we set $n = 100$ and $PMR = 0.3$. In both these scenarios, \ourname is successful in eliminating all the backdoors, obtaining a $BA$ of $0\%$ and achieving an average $MA$ of $92.4\%$ for the CIFAR-10 dataset and $100\%$ for the IoT-Traffic dataset.
    
    \noindent\textbf{Client Order. } To verify that the client updates are exchangeable, 
    we conducted an experiment for the CIFAR-10 dataset, where the models were randomly shuffled. However, the shuffling did not affect the results, as we got $BA=0\%$ and $MA=92.5\%$. These results are intuitive because irrespective of the order in which the client updates arrive at the detection module of \ourname, it does not affect the computation of $Max_{JD}^{i}$, which is eventually used to identify the poisoned updates.

\changed{\noindent\textbf{Adaptive attacks. }\ourname assumes that \adversary knows the backdoor defense deployed at the global server (see~\sect\ref{sec:threat}). Thus, \adversary can constrain the training process to make $H^{t}$ inconspicuous, by
using its benign data to estimate a benign model and thus, $p$ and $H^t$. However, \adversary cannot estimate $q$ as this requires knowing parameters that the server calculates on run-time. Thus, an adaptive attacker can only work with $H^{t}$ to launch backdoor attacks against such defense. In this setting, we conducted experiments for the CIFAR-10,
by updating the loss function of \adversary using the base measure for the anomaly evasion loss term~\cite{bagdasaryan} according to the equation:  
\begin{equation}
    \mathcal{L} = \alpha \mathcal{L_{\text{class}}} + (1-\alpha) \mathcal{L_{\text{BM}}} 
\end{equation}
$\mathcal{L_{\text{class}}}$ captures both the BA and the MA, and $\mathcal{L_{\text{BM}}}$ captures the defense mechanism dependency on the base measure. We conducted three experiments with $\alpha$ (determines the trade-off between model accuracy and evasion from defense mechanism) values as 0.0, 0.5, and 1.0. For $\alpha=0.0$, \adversary sacrifices the model accuracy to evade the defense mechanism, for $\alpha=0.5$, \adversary is equally trading off the model accuracy and defense mechanism evasion, while for $\alpha=1$, \adversary is more concerned about the model accuracy than evading the defense mechanism detection.
For $\alpha=1$, \ourname achieved a $BA$ of $0\%$, $MA$ of $92.33\%$, $TPR=1$ ($TP=10$ and $FN=0$), and $TNR=0.95$ ($TN=19$ and $FP=1$). However, as we decreased $\alpha$ to 0.5, \ourname was effective in detecting and filtering an adaptive attacker's model updates. For $\alpha=0.5$, \ourname obtained a $BA$ of $0\%$, $MA$ of $92.25\%$, $TPR=1$, and $TNR=1$. For $\alpha=0$, \ourname obtained a $BA$ of $0\%$, $MA$ of $92.14\%$, $TPR=0$ ($TP=0$ and $FN=10$), and $TNR=0.85$ ($TN=17$ and $FP=3$). Hence, an adaptive adversary can evade detection at the cost of model accuracy. However, the non-detected models do not have any overall impact on the efficacy of \ourname as the $BA$ is always zero. In summary, our experiments show that \ourname is successful in defending against an adaptive adversary who has working knowledge \mbox{of \ourname deployed at the global server.}
}

\section{Security Analysis}
\label{sec:security}

\vspace{-0.2cm}
\changed{
\noindent This section provides a security analysis to corroborate that \ourname can neutralize backdoors by modeling the defense mechanism using BNP modeling concepts. We explain why our defense works and justify its effectiveness.
To bypass our defense, an adversarial client (\adversary) has to ensure that \ourname cannot distinguish between malicious and benign model updates. Below, we present three mechanisms through which \adversary can hide the backdoors from \ourname. First, \adversary can vary the fraction (PMR) of malicious clients, i.e., \adversary can either reduce the PMR and make the attack less suspicious, or increase the PMR to keep the attack successful while making the models less suspicious. Second, \adversary can limit the poison data rate (PDR) for each adversarial client, i.e., instead of poisoning the entire dataset, \adversary could partially poison the dataset. 
Finally, \adversary can utilize an adaptive attack strategy, such as adding regularization terms (i.e., defense evasion) to the objective function of the training process (see~\sect\ref{subsec:strategies}). A sophisticated \adversary with the working knowledge of \ourname (has access to the previous round base measure $H^{t-1}$) could select a sweet spot between the model accuracy and the evasion from \ourname. As a result, the poisoned models are still similar to the benign models.

In all the above cases, we have demonstrated that \ourname successfully detected all the malicious updates. The reason being \ourname computes an alternate, more generic representation of the client updates, i.e., a probabilistic measure that encompasses all the adjustments made to the client updates due to any local client's training strategy. Hence, the detection module that takes this probabilistic measure as one of its inputs correctly identifies all the malicious updates without being affected by any local client training strategies. In addition, we also integrate the effect of $cos(W_{i}^{t},G^{t-1})$ and \lnorm (Eq.~\ref{eqn:2} and Eq.~\ref{eqn:3}) in the clients' model updates and the computation of error introduced by the client weight. The rationale is that even though \adversary makes sure the distribution of malicious updates does not deviate from benign ones, \adversary cannot fully manipulate the $cos(W_{i}^{t},G^{t-1})$ or \lnorm. The reason being \adversary aims to simulate the global model in the backdoor direction. This ensures that any changes the strategic \adversary makes utilizing advanced hiding techniques cannot bypass \ourname.
We also empirically verified the effectiveness of \ourname using \sota (CIFAR-10, MNIST, and FMNIST) and real-world (IoT) datasets and successfully demonstrated that \adversary cannot conduct backdoor attacks while simultaneously bypassing our defense mechanism. Therefore, \ourname is robust and resilient against backdoor attacks.
}
\section{Related Works}
\label{sec:related}
\vspace{-0.2cm}
\changed{
\noindent Defense mechanisms (against backdoor attacks) in the literature can be broadly classified into two categories: \emph{detection-based} defense mechanisms~\cite{shen16Auror,fung2020FoolsGold,munoz19AFA,diakonikolas2019sever,khazbak2020mlguard,li2019abnormal,li2020learning} and \emph{mitigation-based} defense mechanisms~\cite{yin2018byzantine,guerraoui2018hidden,pillutla2022robust,sun2019can,wu2020mitigating,xie2021crfl}. Detection-based defenses detect and filter the poisoned updates using similarity measures between the poisoned and benign updates. 
In contrast, mitigation-based defenses construct aggregation rules or add noise to the updates to mitigate the poisoned updates which are unbeknown to them. \\
\textbf{Detecting backdoors. }Detection-based defense mechanisms in the literature include: Auror \cite{shen16Auror}, Krum \cite{blanchard17Krum}, AFA \cite{munoz19AFA}, and FoolsGold \cite{fung2020FoolsGold}. However, these defense mechanisms work only when certain conditions are satisfied. For example,  Auror and Krum only work for benign \iid data. In contrast, FoolsGold overcomes this assumption by assuming the benign data is \nonIid and that the manipulated data is \iid. In addition, these defense mechanisms can be bypassed if an adversary restricts the malicious updates within the valid range of benign updates distribution. In summary, these defenses only work when certain conditions are satisfied. On the contrary, \ourname does not assume anything about the distribution of local client's data. \mbox{Thus, it works more effectively against such attacks.}\\
\textbf{Mitigating backdoors. } Mitigation-based defenses include rule-based aggregation mechanisms such as coordinate-wise median and coordinate-wise trimmed mean~\cite{yin2018byzantine}, a two-step aggregation algorithm that combines the Krum and trimmed mean mechanisms~\cite{guerraoui2018hidden}, and RFA~\cite{pillutla2022robust}. These defense mechanisms determine a client update to be benign if it lies within the scope of some aggregation rule. These rules, however, can be easily bypassed if an adversary makes sure its update is within the valid range of these rules. In addition, these rules are computationally intensive. Differential privacy (DP) defense mechanisms~\cite{sun2019can,wu2020mitigating,xie2021crfl,naseri2020local} have also been designed to protect against backdoor attacks. These defense mechanisms follow clipping of the weights and additive noising~\cite{Dwork}, to limit the impact of the adversarial updates. However, they also decrease the $MA$ simultaneously. Nguyen \etal~\cite{nguyen22Flame} designed a defense to limit the impact of noise on $MA$, however, the outlier detection is prone to removing benign models, which reduces the performance in \nonIid scenarios. In comparison, the BNP modeling and CRP-Jensen of \ourname allow us to effectively distinguish between benign and poisoned models.}

\section{Conclusion}
\label{sec:conclusion}
\vspace{-0.2cm}
\changed{
\noindent This paper proposes \ourname, a novel and more generic probabilistic approach to defend against backdoor attacks in Federated Learning. 
In contrast to existing defenses that mainly consider models as a set of vectors and matrices~\cite{blanchard17Krum,fung2020FoolsGold,mcmahan2018iclrClipping,munoz19AFA,nguyen22Flame,shen16Auror} and operate \emph{directly} on them, \ourname first computes a probabilistic measure over the clients' updates that encompass all the adjustments made in the updates due to any local client training strategy. Then, \ourname employs a detection algorithm that utilizes this probabilistic measure to detect and filter out malicious updates. Thus, it overcomes several shortcomings of previous backdoor defense approaches. 
\ourname utilizes two extensions of Bayesian non-parametric modeling techniques: the Hierarchical Beta-Bernoulli Process to draw a probabilistic measure given the clients' model updates (or weights), and a variation of the Chinese Restaurant Process, CRP-Jensen, which is a clustering algorithm that can leverage the probabilistic measure to detect and filter out malicious updates.
Our extensive evaluation with benchmark datasets in different domains demonstrates that \ourname can effectively mitigate backdoor attacks in FL while preserving the \mbox{benign performance of the global model.}
}



\section*{Acknowledgements}
\vspace{-0.2cm}
\noindent This work was funded in part by Intel as part of the Private AI center, HMWK within ATHENE project, Huawei as part of the OpenS3 Lab, the Hessian Ministery of Interior and Sport within the F-LION project, as part of the funding guideline for cyber security research and the US NSF under award number 1943351.
\bibliographystyle{plain}
\bibliography{reference}

\appendices

\section{Federated Learning}
\label{app:fl}
\noindent Federated learning (FL) collaboratively learns a global model $G$ by iteratively aggregating the local model updates sent by the $n$ clients selected during each training round. In each training round $t$, the global model server selects a total of $n$ clients and sends them a common aggregated global model, $G^{t-1}$. Then, each client $i \in \{1,\ldots,n\}$ locally trains a local model $W_{i}^{t}$, using its own local dataset $D_{i}$. After training the local model, each client $i$, sends the updated model parameters to the global server to compute the next stage global model, $G^{t}$.
In this paper, we assume the global server aggregates the local updates by utilizing \emph{Federated Averaging} (\fedavg) function~\cite{mcmahan2017}, given as:
$$G^{t} = \sum_{i=1}^{n} \frac{s_{i}}{s} W_{i}^{t} \text{,} \quad \text{where} \, \, s_{i} = ||D_{i}|| \, \, \text{and} \, \, s = \sum_{i=1}^{n} s_{i}$$

\fedavg utilizes each client's training dataset to compute the latest global model. A malicious client (or clients) can take advantage of this requirement by sending the erroneous dataset size~\cite{wang2020attack}. To eliminate this scenario, we assume equal weights ($s_{i} = 1/n$) for all the clients. Such an approach has also been adopted in previous research efforts~\cite{bagdasaryan,shen16Auror,xie2020dba,rieger2022deepsight}.

\section{Background and Preliminaries}
\label{app:background}
\noindent In this section, we provide brief technical background knowledge on Bayesian non-parametric modeling and related concepts, which will be critical in understanding the design of \ourname.

\subsection{Hierarchical Beta-Bernoulli Process (HBBP)}
\label{sec:background-bernoulli}

\noindent Next, we describe the concepts of the baseline Beta Process, the Hierarchical Beta Process, and the Bernoulli Process, which are used to compute the probabilistic measure.

\noindent
\textbf{Beta Process (BP). }
\changed{A Beta Process is a random discrete measure on the countable infinitely drawn set of weights, where each weight has a mass in the range (0,1) such that the total mass sums to 1.}
A Beta Process uses a concentration function $c$ over some space $\Omega = \mathbb{R}$ and a base measure $H$ to produce some random measure $A$, i.e., $A \sim BP(c, H)$. Given the set $\Omega$, informally, a \emph{measure} is any consistent assignment of \emph{sizes} to (some of) the subsets of the set. Depending on the application, the size of a subset may be interpreted as either its physical size or the probability that some random process will yield a result within the subset. Formally, a measure is a function $\mu:\Sigma \mapsto [0,\infty]$, where $\Sigma$ is the $\sigma$-algebra (collection of subsets) of $\Omega$. 
A concentration function ($c$) quantitatively characterizes the scatter of the values of a random variable. Thus, it indicates the similarity between the input base measure ($H$) and the output random measure ($A$). The base measure $H$ can represent any initial distribution (see~\sect\ref{sec:rationale}). We also call $\gamma_{0} = H(\Omega)$ the mass parameter. So, if $H$ is a normal distribution, then $\gamma_{0}$ is a normal distribution of the complete space $\Omega$. Alternatively, $A$ is a discrete measure (discrete weights), represented by $A = \sum_{j} a_{j} \delta_{w_{j}}$, where $\delta$ is an indicator function. Thus, in order to draw an infinitely countable set of  points ($a_{j}, w_{j}) \to [0,1] \times \Omega$ needs to be drawn. 
The probabilistic weights $\{ a_{j}\}_{j=1}^{\infty}$ are distributed by a stick-breaking process: \mbox{$d_{j} \sim Beta(\gamma_{0}, 1)$}, $a_{j} = \prod_{k=1}^{j} d_{k}$. In a stick-breaking process~\cite{paisley2010stick}, there is a stick of length of 1 and $a_{j}$ represents the probabilistic weight taken from the remainder of the stick every time. $w_{j}$ are drawn independently and identically distributed (IID) from the normalized base measure \mbox{$w_{j}$ $\sim$ $H/H(\Omega)$} with domain $\Omega$. Here, $Beta(\cdot)$ represents the beta distribution that is used to model the continuous random variables in the range [0, 1].  This work assumes that $\Omega$ is simply a space of weights. The objective here is first to draw a baseline Beta prior (random discrete measure) using a Beta Process and then use this prior (in a Hierarchical Beta Process, as explained next) to draw the corresponding Beta priors for the $n$ entities.\\

\noindent
\textbf{Hierarchical Beta Process (HBP).}
\changed{A Hierarchical Beta Process (HBP) is used to create hierarchies of the baseline Beta process when certain conditions are satisfied. Alternatively, from a pool of countable infinite sets of weights of the BP, a subset of weights (under some conditions) are drawn for each sub-Beta Process, creating hierarchies.}
We can employ HBP to draw a discrete random measure corresponding to each of the $n$ entities based on the baseline Beta Process prior~\cite{thibaux2007hierarchical}. 
Let us consider the following rationale for the construction of the HBP.  
Suppose that $W^{Prior}$ is a list of the $n$ entities' weight vectors, 
Now, we assume that the prior for each entity $i$, $W^{Prior}_{i}$ is generated by including weights, which have a specific cosine angular distance, 
with respect to some base weight, $w_{b}$ (different for every entity). Thus, each $W^{Prior}_{i}$ is generated by including $h$ weights ($w$) independently with a probability $p^{h}_{w}$ specific to the entity $i$. These probabilities form a discrete measure $A_{i,h}$ over the space of weights $\Omega$, and we put a Beta Process $BP(c_{i}, A)$ prior on $A_{i,h}$ (Note: $A_{i,h}$ is the same as $A_{i}$ defined in \sect\ref{sec:background} of main paper). In summary, we have the following Hierarchical Beta model:

{
$$\text{Baseline Beta prior:}\quad A \sim BP(c, H)$$
$$ \text{Hierarchical Beta prior:} \quad A_{i,h} \sim BP(c_{i}, A) \quad \forall 1 \leq i \leq n$$
}

The random measure $A$, thus, $A_{i,h}$, encodes the probability that each entity possesses each particular weight.  \\

\noindent
\textbf{Bernoulli Process (BeP). }
\changed{A Bernoulli Process (BeP) is a draw of weights from a space of total weights, given the Beta Process random measure that encodes the probability of selecting the weight in the draw. BeP is employed to draw weights given the Hierarchical Beta priors computed earlier.} Thus, the subsets of points in the HBP prior $A_{i,h}$ are drawn using a BeP with input as the random measure $A_{i,h}$. Each subset $W_{i}$ for entity in $i \in \{1,...,n\}$, having $l$ weights, is characterized by a Bernoulli Process such that $W_{i} | A_{i,h} \sim \text{BeP}(A_{i,h})$. Each subset can also be represented by a discrete measure such that the points ($b_{i,l}, w_{l}$) $\to [0,1] \times \Omega$, forms $W_{i} = \sum_{l} b_{i,l} \delta_{w_{l}}$, where $b_{i,l}$ is the probabilistic weight (success probability) given to $w_{l}$, i.e., \mbox{$Pr(b_{i,l} = 1) = p^{h}_{w}$}, if they are included in the subset $W_{i}$.
\\

\noindent
\textbf{Conjugacy. }It has been shown in the literature that the Beta distribution is the conjugate of the Bernoulli distribution~\cite{broderick2018posteriors}. Hence, we do not have to use the computationally intensive Bayes' rule to compute the posterior distribution of hierarchical random measures. It can be computed as follows: 
Let \mbox{$A_{i,h} \sim BP(c_{i}, A)$}, and let $W_{i} | A_{i,h} \sim \text{BeP}(A_{i,h})$. In $W_{i}= \{W_{i,1}, W_{i,2}, ..., W_{i,l}\}$, $l$ denotes the independent BeP draws over the likelihood function, $A_{i,h}$. By using the results for HBP and BeP in~\cite{kim1999nonparametric}, the posterior distribution of $A_{i,h}$ after observing $W_{i}$ is still a Beta process with modified parameters:
\begin{equation}
    A_{i,h}|W_{i} \sim BP\left(c_{i}+l, \frac{c_{i}}{c_{i}+l} H + \frac{1}{c_{i} \cdot l} \sum_{l = 1}^{l}W_{i} \right) \label{eqn:1}
\end{equation}

\changed{Our motivation is to first draw a baseline Beta Process random measure by drawing a countable infinitely set of weights, such that their probabilistic weight sums to 1. Then, we use this baseline Beta Process to form hierarchies of Beta Process for $n$ different entities. We do so by selecting a subset of $h$ weights from the total weights space for each of the $n$ entities. Then, for each of the $n$ entities, we use Bernoulli Process to draw $l$ weights from the corresponding hierarchical Beta Process weights space. Finally, we keep updating the corresponding hierarchical Beta Process for entity $n$ using the conjugacy of the Beta Process and the Bernoulli Process \cite{broderick2018posteriors}.}

\subsection{Mixture modeling: Chinese Restaurant Process (CRP)}
\label{subsec:crpjensen}
\noindent \changed{The CRP~\cite{socher2011spectral,blei2011distance,li2019tutorial} is a discrete-time stochastic process in the probability theory that resembles the situation of seating customers at tables in a Chinese restaurant with an infinite number of circular tables, each with infinite capacity. The first customer that arrives sits at the first table. The following customers can either sit at the already occupied tables or can choose to sit at the new table. This process partitions the customers among tables. The results of this process are \emph{exchangeable}, meaning that the order in which the customers arrive and sit does not affect the probability of the final distribution. In CRP, we compute two probabilities for table (cluster) assignment. The first probability is the probability of the customer entering the restaurant and sitting at the already occupied tables (clusters). The second probability is the predictive probability of how well this new customer fits the mean of already occupied tables (clusters).}

\changed{\subsection{Jensen-Shannon Divergence}
\noindent Jensen-Shannon Divergence or \jensenDivergence is used to realize the distance between two distributions. Specifically, it is computed by estimating the relative entropy between two distributions. The entropy of a random variable $X$ having probability mass function $P(x)$ is given as:

\begin{equation}
    H(X) = -\sum_{x \in X} P(x)log_{b}P(x)  
\end{equation}

\jensenDivergence is estimated using the Kullback-Leibler divergence (KL-divergence). KL-divergence also measures the distance between two distributions. However, it is not symmetric and does not satisfy the triangle inequality. \jensenDivergence is an approach that improves upon the KL-divergence, as it is symmetric and a smoothed version of KL-divergence. KL divergence between two distributions $p$ and $q$ is given as:

\begin{equation}
    D_{KL}(p||q) = \int_{-\infty}^{\infty} p(x) log \left(\frac{p(x)}{q(x)}\right) dx  
\end{equation}

\jensenDivergence between two distributions $p$ and $q$ is given as:

\begin{equation}
    JSD(p||q) = \frac{1}{2} \left( D_{KL}(p||m) + D_{KL}(q||m) \right)  
\end{equation}

}
\changed{
\section{Overview of Used Symbols}
\label{app:symbols}
\noindent Table~\ref{tab:symbols} contains an overview of the used symbols.
}

\begin{table}
\changed{\caption{Overview of symbols.}
\label{tab:symbols}

\resizebox{\columnwidth}{!}{
\begin{tabular}{l|l}
\textbf{Symbol} & \textbf{Description} \\\hline
BNP & Bayesian non-parametric \\ \hline
HBBP & Hierarchical Beta-Bernoulli Process \\ \hline
BP & Beta Process \\ \hline
HBP & Hierarchical Beta Process \\ \hline
BeP & Bernouli Process \\ \hline
CRP & Chinese Restaurant Process \\ \hline
\textit{f} & Neural Network (NN) \\ \hline
$Beta(\cdot)$ & Beta distribution \\ \hline
$G^{t}$ & Global Model at time $t$ \\ \hline
$W_{i}^{t}$ & Local Model of client i at time $t$ \\ \hline
$D_{i}$ & Local dataset of client $i$ \\ \hline
$n$ & Number of clients \\ \hline
$n_\adversary$ & Number of malicious clients \\ \hline
$t$ & FL training round \\ \hline
$c$ & Concentration function \\ \hline
$\Omega (\mathbb{R})$ & Space of weights \\ \hline
$H$ & Base measure \\ \hline
$A$ & Beta Process random measure \\ \hline
$\gamma_{0}$ & Mass parameter \\ \hline
$\delta$ & Indicator function \\ \hline
$ w_{j}$ & IID weights drawn from $\Omega (\mathbb{R})$ or from BP \\ \hline
$a_{j}$ & Probabilistic weight assigned to $ w_{j}$ \\ \hline
$d_{j}$ & A stick-breaking process \\ \hline
$W^{Prior}$ & List of the $n$ client weight vectors \\ \hline
$W^{Prior}_{i}$ & client $i$'s weight vector \\ \hline
$h$ & Weights drawn from BP to be included in $W^{Prior}_{i}$ \\ \hline
$p^{h}_{w}$ & \begin{tabular}[x]{@{}l@{}}Probability with which $h$ weights are drawn from BP\\ to be included in $W^{Prior}_{i}$\end{tabular} \\ \hline
$A_{i,h}$ or $A_{i}$ & HBP random measure for each client $i$ having weights $h$ \\ \hline
$c_{i}$ &  client $i$'s concentration function \\ \hline
$BP(c_{i}, A)$ & HBP $BP(c_{i}, A)$ prior on $A_{i,h}$ \\ \hline
$w_{l}$ & IID weights drawn from HBP \\ \hline
$b_{i,l}$ & probabilistic weight given to $w_{l}$. Equal to $p_{w}^{h}$ \\ \hline
$D_{KL}(p||q)$ & KL divergence between two distributions $p$ and $q$ \\ \hline
$JSD(p||q)$ & \jensenDivergence between two distributions $p$ and $q$ \\ \hline 
$\mathcal{L}$ & Labels for samples from domain $\mathcal{D}$\\ \hline 
\adversary & Adversary\\ \hline 
$l_\adversary$ & \adversary chosen labels\\ \hline 
$\mathcal{D}_\adversary$ & \textit{trigger set} of \adversary \\ \hline 
$\mathcal{N}$ & Normal distribution \\ \hline 
$\mu_{p}$ & Mean of the flattened initial $G^{t}$ \\ \hline 
$\sigma_{p}$ & Standard deviation of the flattened initial $G^{t}$ \\ \hline 
$W_{i,up}^{t}$ & Updated client weight at time $t$ \\ \hline 
$Max_{JD}^{i}$ & Maximum \jensenDivergence \\ \hline 
$cos(G_{t-1},W_{i}^{t})$ & \begin{tabular}[x]{@{}l@{}}Cosine angular distance between local model $W_{i}^{t}$\\and global model $G_{t-1}$\end{tabular} \\ \hline 
$d_{w_i^t}$ & \lnorm between $G_{t-1}$ and $W_i^t$\\ \hline 
$\sigma_{w_i^t}$ & Measurement error due to the new client's weight \\ \hline 
$\overline{W_{i, up}^{t}}$ & Mean of $W_{i, up}^{t}$ \\ \hline 
$\mu_{c_{l}}$ & Mean of the clusters \\ \hline 
$\sigma_{c_{l}}$ & Variance of the clusters \\ \hline 
$noc$ & Total number of clusters formed yet \\ \hline 
$p_{i}$ & $p$ distribution of client $i$ \\ \hline 
$q_{noc}$ & $noc$ cluster $q$ distribution \\ \hline 
$js_{noc}^{i}$ & \begin{tabular}[x]{@{}c@{}}\jensenDivergence of $noc$ cluster $q$ distribution with\\ $p$ distribution of client $i$\end{tabular} \\ \hline 
$\mu_{new}$ & Updated cluster's mean \\ \hline 
$\sigma_{new}$ & Updated cluster's standard deviation \\ \hline 
$n_{k}$ & \begin{tabular}[x]{@{}l@{}}Number of clients update already assigned to a\\ particular cluster\end{tabular} \\ \hline 
$\tau_{k}$ & Precision of the cluster \\ \hline 
$\mu_{0}$ & Initial mean for the new cluster \\ \hline 
$\tau_{0}$ & Initial Precision assumed for the new cluster \\ \hline 
$Max_{JD}^{stored}=[]$ & Array of stored $Max_{JD}^{i}$ \\ \hline 
\end{tabular}}}
\end{table}

\changed{
\section{\ourname statistics for FMNIST}
\label{app:fmnist}

\begin{figure*}[htbp]
     \centering
	\begin{subfigure}[b]{0.24\textwidth}
         \centering
         \includegraphics[width=\textwidth, trim=0.2cm 0.4cm 0.2cm 0]{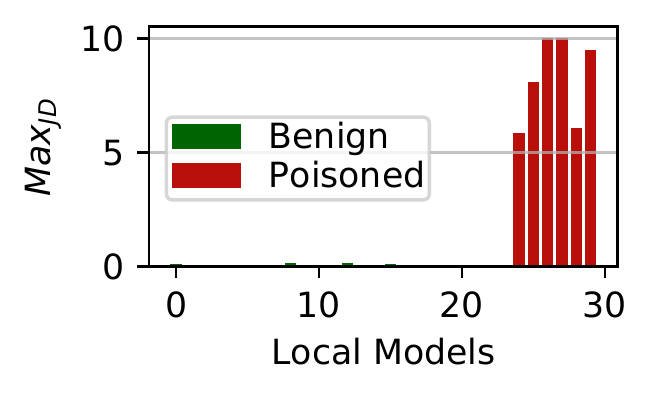}
         \caption{\nonIid = 0.0}
  	    \label{fig:fmnist_iid_0}
    \end{subfigure}
    \begin{subfigure}[b]{0.24\textwidth}
         \centering
         \includegraphics[width=\textwidth, trim=0.2cm 0.4cm 0.2cm 0]{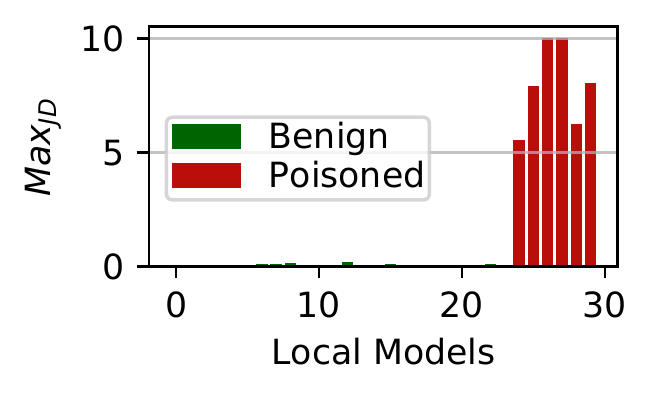}
         \caption{\nonIid = 0.5}
         \label{fig:fmnist_iid_5}
    \end{subfigure}
    \begin{subfigure}[b]{0.24\textwidth}
         \centering
         \includegraphics[width=\textwidth, trim=0.2cm 0.4cm 0.2cm 0]{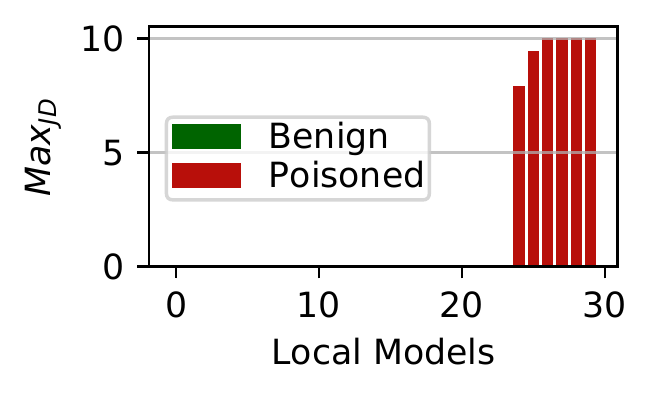}
         \caption{\nonIid = 1.0}
         \label{fig:fmnist_iid_10}
    \end{subfigure}
    \caption{Effect of different \nonIid rates on the maximum \jensenDivergence ($Max_{JD}^{i}$) for FMNIST dataset.}
        \label{fig:fmnist_iid}
\end{figure*}

\begin{figure*}[htbp]
     \centering
	\begin{subfigure}[b]{0.24\textwidth}
         \centering
         \includegraphics[width=\textwidth, trim=0.4cm 0.4cm 0.4cm 0]{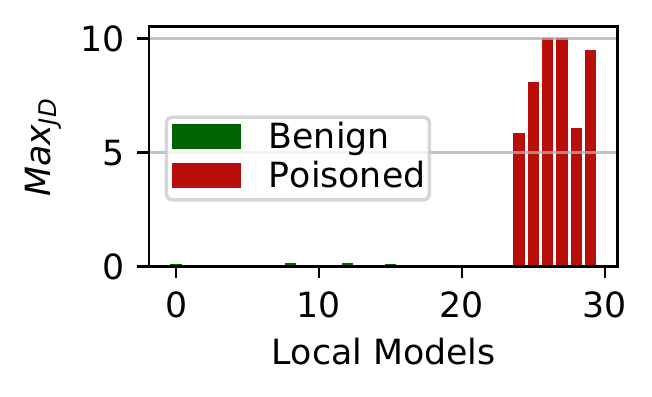}
         \caption{PMR = $20\%$}
  	\label{fig:fmnist_pmr_6}
     \end{subfigure}
      \begin{subfigure}[b]{0.24\textwidth}
         \centering
         \includegraphics[width=\textwidth, trim=0.4cm 0.4cm 0.4cm 0]{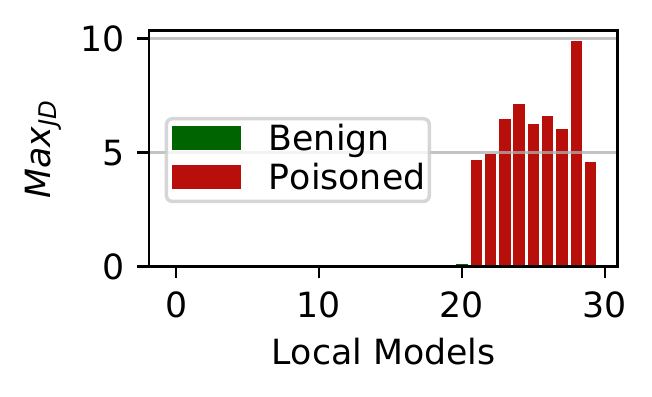}
         \caption{PMR = $30\%$}
         \label{fig:fmnist_pmr_9}
     \end{subfigure}
      \begin{subfigure}[b]{0.24\textwidth}
         \centering
         \includegraphics[width=\textwidth, trim=0.4cm 0.4cm 0.4cm 0]{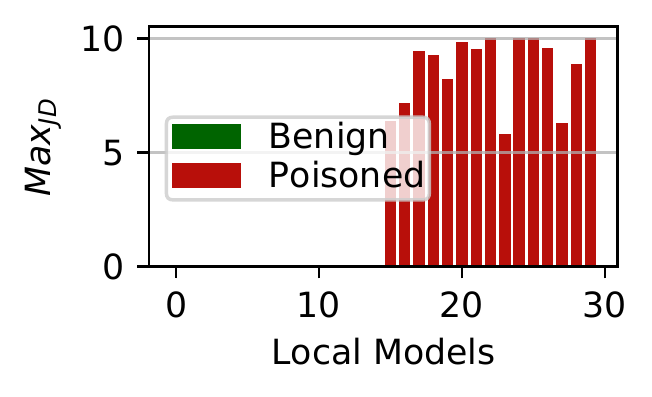}
         \caption{PMR = $50\%$}
         \label{fig:fmnist_pmr_15}
     \end{subfigure}
        \caption{Effect of different PMR rates on the maximum \jensenDivergence ($Max_{JD}^{i}$) for the FMNIST dataset.}
        \label{fig:fmnist_pmr}\vspace{-0.3cm}
\end{figure*}

\begin{figure*}[htbp]
     \centering
	\begin{subfigure}[b]{0.24\textwidth}
         \centering
         \includegraphics[width=\textwidth, trim=0.2cm 0.4cm 0.2cm 0]{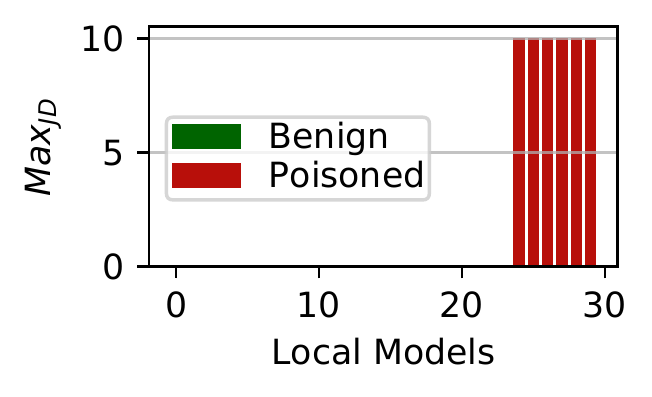}
         \caption{\nonIid = 0.0}
  	    \label{fig:mnistr_iid_0}
    \end{subfigure}
    \begin{subfigure}[b]{0.24\textwidth}
         \centering
         \includegraphics[width=\textwidth, trim=0.2cm 0.4cm 0.2cm 0]{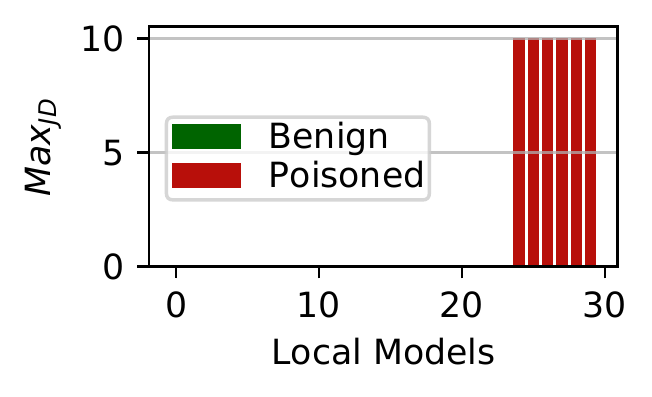}
         \caption{\nonIid = 0.5}
         \label{fig:mnist_iid_5}
    \end{subfigure}
    \begin{subfigure}[b]{0.24\textwidth}
         \centering
         \includegraphics[width=\textwidth, trim=0.2cm 0.4cm 0.2cm 0]{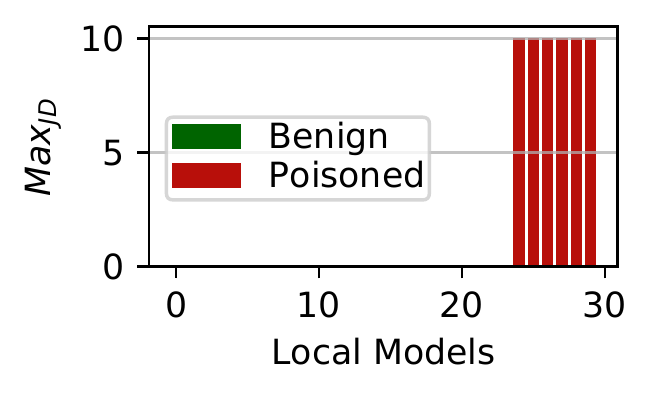}
         \caption{\nonIid = 1.0}
         \label{fig:mnist_iid_10}
    \end{subfigure}
    \caption{Effect of different \nonIid rates on the maximum \jensenDivergence ($Max_{JD}^{i}$) for MNIST dataset.}
        \label{fig:mnist_iid}
\end{figure*}

\begin{figure*}[htbp]
     \centering
	\begin{subfigure}[b]{0.24\textwidth}
         \centering
         \includegraphics[width=\textwidth, trim=0.4cm 0.4cm 0.05cm 0]{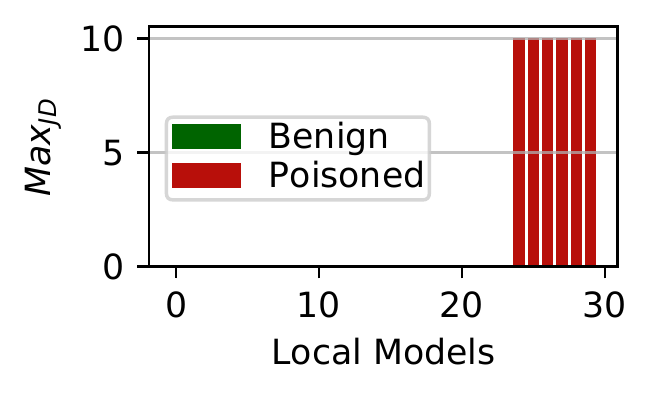}
         \caption{PMR = $20\%$}
  	\label{fig:mnist_pmr_6}
     \end{subfigure}
      \begin{subfigure}[b]{0.24\textwidth}
         \centering
         \includegraphics[width=\textwidth, trim=0.4cm 0.4cm 0.05cm 0]{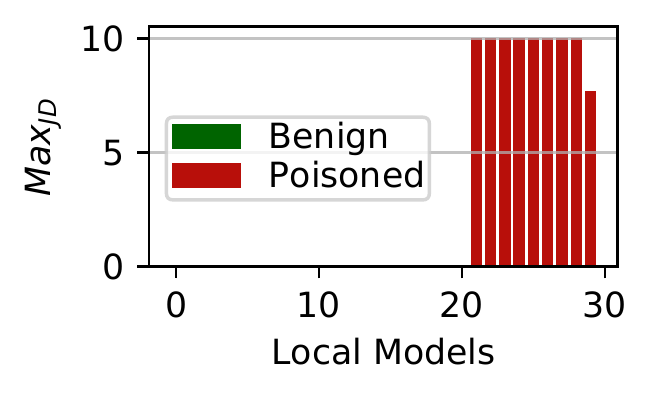}
         \caption{PMR = $30\%$}
         \label{fig:mnist_pmr_9}
     \end{subfigure}
      \begin{subfigure}[b]{0.24\textwidth}
         \centering
         \includegraphics[width=\textwidth, trim=0.4cm 0.4cm 0.05cm 0]{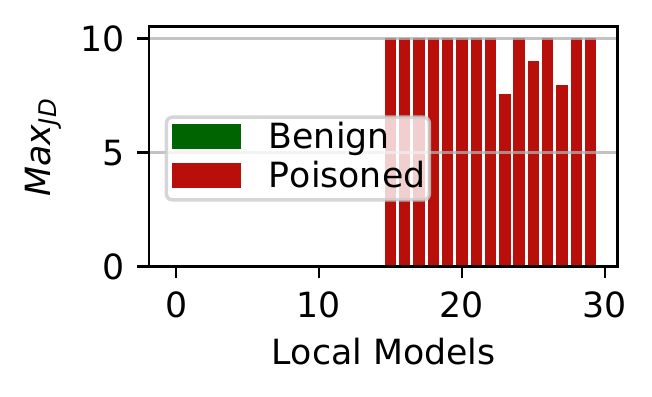}
         \caption{PMR = $50\%$}
         \label{fig:mnist_pmr_15}
     \end{subfigure}
        \caption{Effect of different PMR rates on the maximum \jensenDivergence ($Max_{JD}^{i}$) for the MNIST dataset.}
        \label{fig:mnist_pmr}\vspace{-0.3cm}
\end{figure*}

\noindent In this section, we evaluate the impact of \nonIid rate and $PMR$ on the FMNIST dataset. First, we demonstrate the trend of $Max_{JD}^{i}$ for both the malicious and benign clients with respect to each of these parameters. Then, we illustrate the impact of \nonIid rate and $PMR$ on \ourname's performance by quantifying different metrics as stated in \sect\ref{sec:setup} and also compare it against the no defense scenario. 

\noindent
\textbf{Illustration of $Max_{JD}^{i}$. }
The impact of the degree of \nonIid data and $PMR$ on $Max_{JD}^{i}$ for each client is shown in Fig.~\ref{fig:fmnist_iid} and Fig.~\ref{fig:fmnist_pmr}, respectively. We select a total of 30 (\numberOfClients = 30) clients for both the \nonIid and $PMR$ experimental analysis. For \nonIid analysis, we test \nonIid $\in \{0.0, 0.5, 1.0\}$ and set $PMR=0.2$. Thus, the number of malicious clients equals 6 (\numberOfMaliciousClients = 6). For $PMR$ analysis, we test $PMR \in \{0.2, 0.3, 0.5\}$, i.e., when $\numberOfMaliciousClients$ equals 6, 9, and 15 and set \nonIid = 0.7. As illustrated in Fig.~\ref{fig:fmnist_iid} and Fig.~\ref{fig:fmnist_pmr}, the $Max_{JD}^{i}$ value for benign clients differs significantly from that of malicious clients. Hence, \ourname easily filters out all the malicious client updates, achieving a $BA$ of zero while keeping the $MA$ of the global model intact.}

\changed{
\section{\ourname statistics for MNIST}
\label{app:mnist}

\noindent In this section, we evaluate the impact of \nonIid rate and $PMR$ on the MNIST dataset. First, we demonstrate the trend of $Max_{JD}^{i}$ for both the malicious and benign clients, with respect to each of these parameters. Then, we illustrate the impact of \nonIid rate and $PMR$ on \ourname's performance by quantifying different metrics as stated in \sect\ref{sec:setup} and also compare it against the no defense scenario. 

\noindent
\textbf{Illustration of $Max_{JD}^{i}$. }
Impact of the degree of \nonIid data and $PMR$ on $Max_{JD}^{i}$ for each client is shown in Fig.~\ref{fig:mnist_iid} and Fig.~\ref{fig:mnist_pmr}, respectively. We select a total of 30 (\numberOfClients = 30) clients for both the \nonIid and $PMR$ experimental analysis. For \nonIid analysis, we test \nonIid $\in \{0.0, 0.5, 1.0\}$ and set $PMR=0.2$. Thus, number of malicious clients equals 6 (\numberOfMaliciousClients = 6). For $PMR$ analysis, we test $PMR \in \{0.2, 0.3, 0.5\}$, i.e., when $\numberOfMaliciousClients$ equals 6, 9, and 15 and set \nonIid = 0.7. As illustrated in Fig.~\ref{fig:mnist_iid} and Fig.~\ref{fig:mnist_pmr}, the $Max_{JD}^{i}$ value for benign clients differs significantly from that of malicious clients. Hence, \ourname easily filters out all the malicious client updates, achieving a $BA$ of zero while keeping the $MA$ of the global model intact.}

\section{Additional adaptive attack}
\label{app:adaptive}
To implement an adaptive attack in which an adversary makes small changes to client updates to keep the JD divergences small, we vary the Poisoned Data Rate (PDR) to demonstrate the increment in the client updates for the CIFAR-10 dataset. We choose PDR to implement such an adaptive attack because arbitrary increments in PDR will also reflect the random increments in the client updates. Then, we compute the $TPR$, $TNR$, $BA$, and $MA$ metrics to evaluate the effectiveness of \ourname compared against the no-defense $BA$ and $MA$, for different PDR values. We conduct experiments for three cases: a) PDR $\in$ (0.1,0.75) with an increment of 0.1, b) PDR $\in$ (0.01,0.1) with an increment of 0.01, and c) PDR $\in$ (0.1,0.2) with an increment of 0.01. Case a) demonstrates the impact of large PDR increments on the metrics mentioned above. The $BA$ with no defense remains at zero for the initial PDR values of 0.01 and 0.1, and after that, it starts to increase. \ourname easily identified all the poisoned updates, thus filtering out all the poisoned updates, achieving a $BA$ of zero while keeping the $MA$ of the global model intact. As we observed in case a), the no-defense $BA$ remains zero at PDR = 0.1 and starts to increase after that; we conducted two more experiments for PDR $\in$ (0,0.1) and PDR $\in$ (0.1,0.2) to analyze \ournameGen performance when the PDR increases. Case b) demonstrates the impact of very small increments in PDR, and the no defense $BA$ remains at zero in this case. Nevertheless, \ourname was correctly able to identify all the poisoned updates. In the case of c), the no defense $BA$ starts to increase from PDR = 0.11, and \ourname again correctly identified all the poisoned updates, thus achieving a $BA$ of zero while keeping the $MA$ of the global model intact. This experimental analysis demonstrates that even when a shrewd adversary makes small iterative changes to PDR (in consequence, client updates), \ourname works efficiently by identifying all the poisoned updates.

\end{document}